\documentclass[10pt,journal,compsoc]{IEEEtran}
\usepackage{color}
\usepackage{xspace}
\usepackage{amsfonts}
\usepackage{amsmath}
\usepackage{amssymb}
\usepackage{bm}
\usepackage{booktabs}
\usepackage{multirow}
\usepackage{graphicx}
\usepackage{threeparttable}
\usepackage{acronym}
\usepackage{makecell}
\usepackage[dvipsnames]{xcolor} 

\usepackage{colortbl}
\definecolor{graycolor}{rgb}{0.95,0.95,0.95}
\usepackage{pifont}
\usepackage[caption=false]{subfig}
\makeatletter
\DeclareRobustCommand\onedot{\futurelet\@let@token\@onedot}
\def\@onedot{\ifx\@let@token.\else.\null\fi\xspace}

\makeatother

\ifCLASSOPTIONcompsoc
  \usepackage[nocompress]{cite}
\else
  \usepackage{cite}
\fi
\ifCLASSINFOpdf
\else
\fi

\begin{document}
\title{Flow-Anything: Learning Real-World Optical Flow Estimation from Large-Scale Single-view Images}

\author{Yingping~Liang,
        Ying~Fu,~\IEEEmembership{Senior~Member,~IEEE},
        Yutao~Hu, Wenqi Shao, Jiaming~Liu, 
        and~Debing~Zhang
\IEEEcompsocitemizethanks{
    \IEEEcompsocthanksitem Yingping Liang and Ying Fu are with School of Computer Science and Technology, Beijing Institute of Technology ({liangyingping,
    fuying}@bit.edu.cn).
    \IEEEcompsocthanksitem Yutao Hu is with School of Computer Science and Engineering, Southeast University (huyutao@seu.edu.cn).
    \IEEEcompsocthanksitem Wenqi Shao is with Shanghai Artificial Intelligence Laboratory (weqish@link.cuhk.edu.hk).
    \IEEEcompsocthanksitem Jiaming Liu is with Tiamat AI (jmliu1217@gmail.com).
    \IEEEcompsocthanksitem Debing Zhang is with Xiaohongshu (debingzhangchina@gmail.com).
    \protect\\
    }
}

\markboth{Journal of \LaTeX\ Class Files,~Vol.~14, No.~8, August~2015}%
{Shell \MakeLowercase{\textit{et al.}}: Bare Demo of IEEEtran.cls for Computer Society Journals}

\IEEEtitleabstractindextext{%
\begin{abstract}

Optical flow estimation is a crucial subfield of computer vision, serving as a foundation for video tasks. However, the real-world robustness is limited by animated synthetic datasets for training. This introduces domain gaps when applied to real-world applications and limits the benefits of scaling up datasets.
To address these challenges, we propose \textbf{Flow-Anything}, a large-scale data generation framework designed to learn optical flow estimation from any single-view images in the real world. We employ two effective steps to make data scaling-up promising. First, we convert a single-view image into a 3D representation using advanced monocular depth estimation networks. This allows us to render optical flow and novel view images under a virtual camera. Second, we develop an Object-Independent Volume Rendering module and a Depth-Aware Inpainting module to model the dynamic objects in the 3D representation. These two steps allow us to generate realistic datasets for training from large-scale single-view images, namely \textbf{FA-Flow Dataset}. For the first time, we demonstrate the benefits of generating optical flow training data from large-scale real-world images, outperforming the most advanced unsupervised methods and supervised methods on synthetic datasets. Moreover, our models serve as a foundation model and enhance the performance of various downstream video tasks.

\end{abstract}

\begin{IEEEkeywords}
Optical flow estimation, novel-view synthesis, warping, stereo matching, and unsupervised learning.
\end{IEEEkeywords}}

\maketitle

\IEEEdisplaynontitleabstractindextext

%
\IEEEpeerreviewmaketitle

\IEEEraisesectionheading{\section{Introduction}\label{sec:introduction}}

\IEEEPARstart{O}{ptical} flow is the precise calculation of per-pixel motion between consecutive video frames. Its applications span a wide range of fields, including object tracking \cite{8735957, liu2024siamese}, robot navigation \cite{karoly2019optical}, three-dimensional (3D) reconstruction \cite{kokkinos2021learning}, and visual simultaneous localization and mapping (SLAM) \cite{zhang2020flowfusion, cheng2019improving}. Therefore, optical flow estimation is a fundamental problem and also requires a foundation model with strong zero-/few-shot performance to estimate motion information from two frames in the real world. Recently, driven by the rapid advancement of deep learning and neural networks, learning-based methods \cite{sun2018pwc, teed2020raft, 9018073, 6104059} have demonstrated significant advances compared to traditional model-based algorithms \cite{weinzaepfel2013deepflow}. Conventional practices primarily rely on synthetic datasets, as shown in \cite{dosovitskiy2015flownet, ilg2017flownet}. Synthetic data typically includes precise dense optical flow labels and animated images utilizing rendering engine. However, the difference and domain gap between synthetic data and real-world scenarios limits its effectiveness and robustness in real-world applications.

Recent efforts to extract sparse optical flow from real-world data have utilized custom-made specialized hardware \cite{geiger2012we, menze2015object, fuying-2013-3DIMPVT, 10310261, 10419040}. However, these methods are hindered by the rigid and inefficient data collection procedures, limiting their real-world use \cite{de2021neural, chen2023instance}. To overcome this challenge, Depthstillation \cite{aleotti2021learning} and RealFlow \cite{han2022realflow} are introduced. These methods involve projecting each pixel of real-world images onto novel view frames using random virtual camera motions or estimated flows. Despite these innovations, the lack of image realism in these methods results in issues like collisions, holes, and artifacts, which negatively impact the performance of learning-based optical flow models \cite{sun2018pwc, teed2020raft}. Thus, the field remains underdeveloped due to the challenges in creating large-scale, realistic datasets with accurate and dense optical flow labels.

In this work, we aim to develop a robust foundational model for optical flow estimation that can generate high-quality optical flow information for real-world images. This requires realistic, large-scale optical flow data for training. To enhance image realism in generating such data, we focus on single-view Multiplane Images (MPI) \cite{single_view_mpi, han2022single}. This method has shown exceptional single-view image rendering capabilities and effectively reduces issues like collisions, holes, and artifacts observed in earlier methods \cite{aleotti2021learning, han2022realflow}. These improvements in image realism raise an important question: Can high-realistic MPI methods for novel view synthesis be adapted to create high-quality optical flow datasets for training purposes?

To this end, we propose \textbf{Flow-Anything}, a method aimed at generating realistic optical flow datasets from single-view images in the real world. First, we carefully analyze the image synthesis pipeline of MPI and develop a framework for generating optical flow alongside novel view images. Using layered planes, we extract optical flow from both rendered and real images by applying virtual camera motions. In this process, we warp features of single-view images onto each layered plane and construct an MPI. A neural network is used to predict the RGB color, density, and optical flow in each plane. These attributes are then mapped into novel view images and optical flow maps via the volume rendering technique.

Second, as MPI is limited to static backgrounds, it contains only plain optical flow with similar motion directions and thus restricts motion realism. To solve this problem, we introduce an Object-Independent Volume Rendering module and a Depth-Aware Inpainting module. The Object-Independent Volume Rendering module separates dynamic objects from static backgrounds and applies different virtual camera motions to calculate the motion of both dynamic and static components. In this way, the dynamic objects have different directions of motion than the static background. Then, to handle object occlusion due to dynamic object motion in the synthesized novel view image, we propose the Depth-Aware Inpainting module. This module uses the depths of novel views obtained through volume rendering to identify occlusion and then perform inpainting.

With Flow-Anything, we can leverage a large number of single-view images from the real world to create training datasets featuring realistic images and motions. Therefore, we have collected a large hybrid dataset to provide a rich image source for data synthesis. Using this dataset, we synthesize the largest known optical flow dataset with real-world images, namely \textbf{FA-Flow Dataset}. The generated dataset contains diverse images from various scenes. This enhances the ability of learning-based optical flow estimation models to generalize across a wide variety of real-world scenes. Extensive experiments on real-world optical flow datasets and various downstream video tasks validate the effectiveness and robustness of our method. In summary, our main contributions are as follows:

\begin{enumerate}
\item We introduce a novel optical flow training data generation framework using single-view real-world images, namely Flow-Anything, which uses multi-plane images as 3D representation and can provide realistic optical flow data for training. 
\item To model dynamic objects and handle occlusion in static single-view images, we present an Object-Independent Volume Rendering module and a Depth-Aware Inpainting module to model dynamic objects in MPI, to separate dynamic objects and fill the holes created by their motion. By utilizing these two modules, realistic optical flow and novel view images can be rendered in dynamic scenes.
\item To our knowledge, this is the first attempt at generating large-scale optical flow data from the real world and training. The experiments show that our training scheme is significantly superior to the existing state-of-the-art unsupervised methods and supervised training schemes using synthetic datasets.
\end{enumerate}

\begin{figure*}[t]
\begin{center}
    \includegraphics[width=1.0\linewidth]{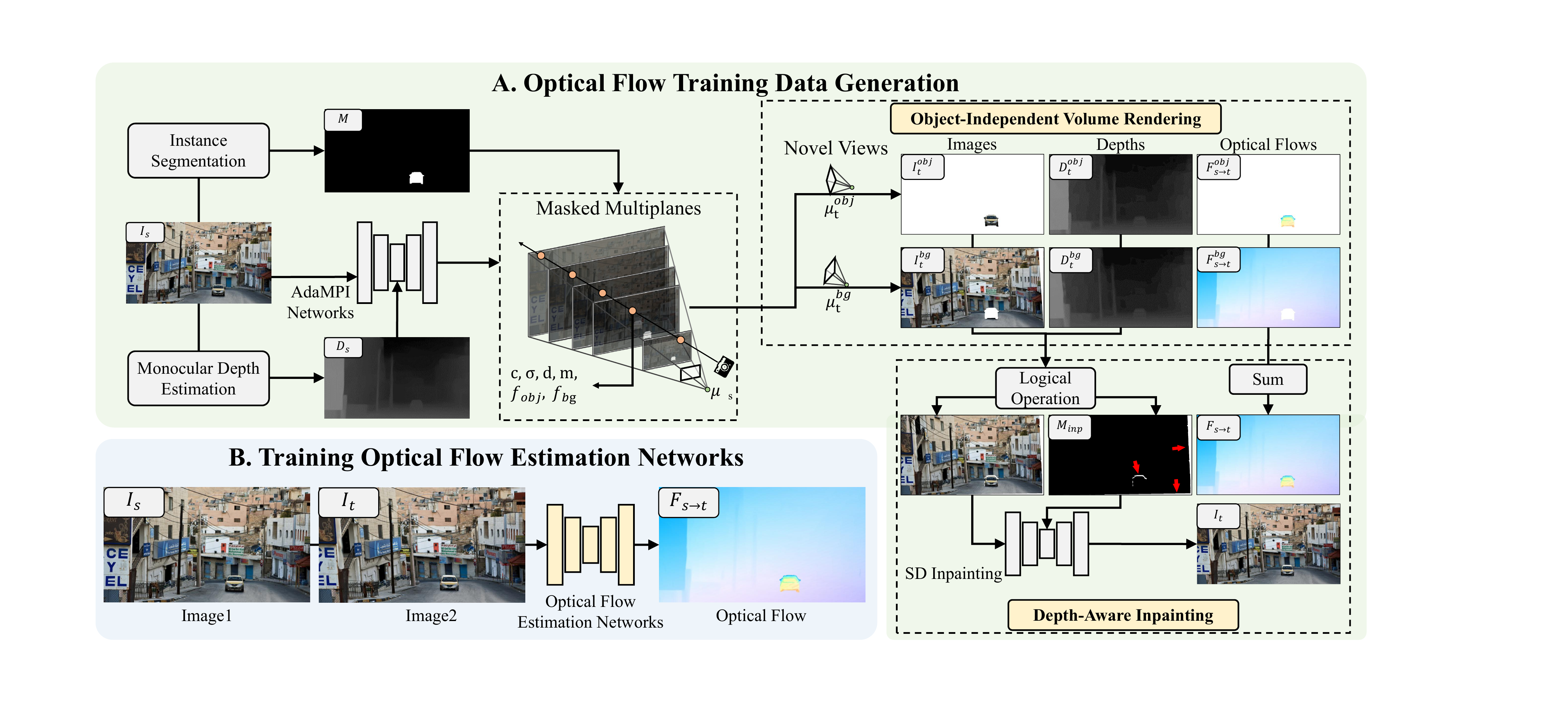}
    \end{center}
    \caption{Illustration of our proposed Flow-Anything, with two main parts. A) Optical flow training data generation pipeline. We estimate depths to construct MPI, where neural networks predict RGB and density for each plane, and virtual cameras calculate the flow. Novel views and flow maps are separated using object masks. Then, the Object-Independent Volume Rendering module is used to render the novel views and optical flows. Holes due to object occlusion are filled with the Depth-Aware Inpainting module. B) Training optical flow estimation networks. The generated training data, including two images and the optical flow, is used to train optical flow networks.}
    \label{fig:framework-v2}
\end{figure*}

\section{Related Work}
\label{sec:related}
In this section, we first introduce the existing optical flow datasets and how to get optical flow labels. Then, we review the most relevant studies on supervised optical flow estimation networks, optical flow dataset generation, and novel view image synthesis methods.

\subsection{Optical Flow Datasets} 

Recent advancements in Optical flow estimation are largely driven by the availability of synthetic datasets. Traditional datasets, such as FlyingChairs \cite{dosovitskiy2015flownet}, FlyingThings3D \cite{ilg2017flownet}, Sintel \cite{butler2012naturalistic}, Virtual KITTI 2 \cite{cabon2020virtual}, and Spring \cite{Mehl2023_Spring}, have played a crucial role in the development and benchmarking of optical flow algorithms. These datasets, however, often consist of synthetic images or images with substantial post-processing, which can introduce a domain gap when applying these models to real-world applications. Real-world datasets, such as HD1K \cite{kondermann2015stereo} and KITTI 12/15 \cite{geiger2012we, menze2015object}, have attempted to bridge this gap by providing data from actual scenes. However, these datasets primarily offer sparse annotations and contain a relatively small number of images, which can restrict the training process for more complex models \cite{zheng2024multi}. To solve this problem, our work focuses on generating data from the real world.

\subsection{Supervised Optical Flow Estimation Networks}
Supervised optical flow estimation networks are usually trained deep neural networks to match patches across images \cite{weinzaepfel2013deepflow}. FlowNet \cite{dosovitskiy2015flownet} is the first neural network with convolutional layers, which is trained on synthetic datasets with dense optical flow labels. Subsequent methods \cite{hui2018liteflownet, hur2019iterative, ilg2017flownet, luo2022learning, liu2024transformer} introduce advanced modules and improves the performance of optical flow estimation by supervised learning. RAFT \cite{teed2020raft}, FlowFormer++ \cite{shi2023flowformer++}, and SEA-RAFT \cite{wang2024sea} are the representative state-of-the-art networks with advanced network design in optical flow estimation. However, the domain gap between real-world applications and synthetic datasets remains a concern, limiting generalization. Consequently, many methods focus on synthetic scenarios similar to their training sets, neglecting real-world performance. Our work aims to address this gap by generating realistic optical flow datasets from real-world images to train such supervised optical flow estimation networks.

\subsection{Optical Flow Dataset Generation}

The data generation method is designed to compensate for the limitations of real-world data \cite{wei2021physics, liang2025relation, zhang2022guided}. Real-world optical flow acquisition is difficult and often requires special hardware \cite{fu2022low, zou2024eventhdr}. The luorescent texture is first utilized to record motions in real-world scenes, which serves as optical flow labels as described in \cite{baker2011database}. In addition, KITTI \cite{geiger2012we, menze2015object} provides sparse optical flow through complex camera setups and expensive hardwares to obtain the corresponding depth of camera settings and calculate optical flow. 
However, these real-world datasets have limited quantities with constrained scenes, which makes it challenging for models to be generalized to broader scenes. As a result, the models may perform poorly in varied real-world scenarios. This highlights the need for more comprehensive datasets. Synthetic optical flow datasets have shown promise using advanced rendering engines, in which FlyingChairs \cite{dosovitskiy2015flownet} and FlyingThings \cite{ilg2017flownet} are representative and widely used datasets. Sintel and Virtual KITTI are also photo-realistic synthetic datasets that use the Unity Engine. However, these synthetic datasets usually fail to accurately replicate real-world scenes, leading to domain gaps. 

AutoFlow~\cite{sun2021autoflow} introduces a learning-based method for generating training data by hyperparameter searching but also relies on optical flow labels for domain adaptation. Two recent works, Depthstillation~\cite{aleotti2021learning} and MPI-Flow~\cite{liang2023mpi}, synthesize paired images and optical flow by estimating depth from a single image and computing flow based on virtual camera poses. RealFlow~\cite{han2022realflow} synthesizes intermediate frames using estimated flow from pre-trained models, leveraging advances in video frame interpolation. Our goal is to scale up optical flow dataset generation from easily accessible single-view images. Therefore, our method builds upon the core idea of MPI-based rendering introduced in MPI-Flow and introduces several improvements. Specifically, we integrate a more accurate depth estimation network to enhance geometric consistency, and employ a generative inpainting model based on Stable Diffusion to robustly fill occlusions and vacant regions arising from novel view synthesis.

\subsection{Novel View Synthesis}
Novel view synthesis methods aim to generate novel views of complex scenes by optimizing an underlying scene function using a sparse set of input images with camera poses. Early methods \cite{lombardi2019neural, mildenhall2021nerf, Zhang_2022_CVPR} utilize multiple views of a scene to render novel views with geometric consistency. Synthesizing novel views from a single image is more challenging due to limited scene information. Geometry-Free View Synthesis \cite{rombach2021geometry} and PixelSynth \cite{rockwell2021pixelsynth} address single-view view synthesis by training neural networks with multi-view supervision. However, they struggle to generalize to more complex scenes due to the lack of large-scale datasets with multi-view images and calibrated camera poses for training. Recent methods decompose scenes into multiple layers and show promising generalization. For example, Single-view MPI \cite{single_view_mpi} predicts disoccluded content and introduces scale-invariant view synthesis for supervision. MINE \cite{li2021mine} introduces the neural radiance fields, which jointly reconstruct the scene function and fill in occluded contents. Then, AdaMPI \cite{han2022single} proposes a novel plane adjustment network to handle complex 3D scene structures. These MPI-based methods have shown success, which we choose as our basic tool for novel view synthesis. However, there are no publicly available methods that currently generate optical flow datasets from MPI. To extract optical flow from MPI, we propose a novel pipeline. This pipeline uses layered depths and virtual camera poses. Besides, we introduce an Object-Independent Volume Rendering module for decoupling static background and dynamic objects, which is typically overlooked in these methods. We also introduce a Depth-Aware Inpainting module to address unnatural occlusions caused by dynamic object motion.

\section{Method}
\label{sec:method}

In this section, we first introduce our motivation and the basic formulation for novel view image generation. Then, we detail the MPI construction pipeline. We also present two crucial components of our method, including the Object-Independent Volume Rendering module and the Depth-Aware Inpainting module. Finally, we introduce the process of training optical flow networks. The illustration of our framework is shown in Figure~\ref{fig:framework-v2}.

\subsection{Motivation and Formulation}

Our goal is to generate a realistic novel view image $\textbf{I}_t\in\mathbb{R}^{H\times W\times 3}$ and the corresponding optical flow maps $\textbf{F}_{s\rightarrow t}\in\mathbb{R}^{H\times W\times 2}$ from a single-view image $\textbf{I}_s\in\mathbb{R}^{H\times W\times 3}$. $H$ and $W$ are the height and width of the image, respectively. The two-dimensional array of the optical flow $\textbf{F}_{s\rightarrow t}$ represents the displacement of corresponding pixels from image $\textbf{I}_s$ to image $\textbf{I}_t$. The input image, generated novel view image, and optical flow together form a data pair for training. 

To generate training data, previous works \cite{aleotti2021learning, han2022realflow} warp pixels from image $\textbf{I}_s$ to image $\textbf{I}_t$ with estimated flows. This leads to holes and artifacts in the image $\textbf{I}_t$, which damages image realism. Recent works \cite{single_view_mpi, li2021mine, han2022single} on Multiplane Images (MPI) reveal that the layered depth representation of a single-view image can significantly improve the realism of the generated novel view image. Therefore, MPI-Flow~\cite{liang2023mpi} recently explores using MPI to synthesize training data for optical flow, but it is limited to small-scale experiments.

In this work, we focus on scaling up this approach to generate high-quality optical flow data from a wide range of single-view images. To this end, we present \textbf{Flow-Anything}, an updated MPI-based optical flow dataset generation framework designed for large-scale and diverse data synthesis. As illustrated in Figure~\ref{fig:framework-v2}, our pipeline incorporates advanced components such as object-aware rendering, depth-aware fusion, and generative inpainting, enabling the training of state-of-the-art optical flow networks at scale.



Besides, due to viewpoint changes and occlusions, merging these novel view images can introduce vacant regions and false occlusions. To address this, we propose a Depth-Aware Inpainting module. Guided by the rendered depth and object masks, we use a generative model based on Stable Diffusion to fill missing regions and correct occlusions in the synthesized image $\textbf{I}_t$.

Finally, the generated image pair $(\textbf{I}_s, \textbf{I}_t)$ and the corresponding optical flow $\textbf{F}_{s\rightarrow t}$ are used to train optical flow networks, which take $\textbf{I}_s$ and $\textbf{I}_t$ as input and are supervised by the generated flow.

\subsection{MPI Construction}
\label{sec3-2}

Following MPI-Flow~\cite{liang2023mpi} and AdaMPI~\cite{han2022single}, we adopt a similar strategy for Multiplane Image (MPI) construction. Specifically, our neural network $\mathcal{F}$ is designed to generate $N$ fronto-parallel RGB$\sigma$ planes from the source view $\bm{\mu}_s$ each with a color map $\textbf{C}_n$, a density map $\bm{\Sigma}_n$, and a depth map $\textbf{D}_n$. The density and depth maps are derived from the monocular depth map $\textbf{D}_s$, which is estimated using a state-of-the-art monocular depth estimation model~\cite{depthanything}.

By discretizing the continuous depth range into $N$ uniformly sampled planes, our method effectively decomposes the scene into layered representations at different depths. Then, the network simultaneously maps image features from $\textbf{I}_s$ to each plane to assign color maps, completing the initial construction of the MPI:

\begin{equation}
    \{(\textbf{C}_n, \bm{\Sigma}_n, \textbf{D}_n)\}^N_{n=1} = \mathcal{F}(\textbf{I}_s, \textbf{D}_{s}),
    \label{eq:eq1}
\end{equation}
where we set $N$ to 64 by default.

To also get the corresponding optical flow labels, we add an additional optical channel in each plane. To this end, each pixel contains the value of $(c, \sigma, d, f_{obj}. f_{bg})$, where $f_{obj}$ and $f_{bg}$ are the pixel shift vectors from source view $\bm{\mu}_{s}$ to target views $\bm{\mu}_{t}^{obj}$ and $\bm{\mu}_{t}^{bg}$, respectively. 

For object motion under view $\bm{\mu}_{t}^{obj}$, we compute the optical flow map on the $n$-th plane at pixel $[x_s, y_s]$ of source image $\textbf{I}_s$ by $\textbf{F}_{n}^{obj}=[x_t^{obj}-x_s, y_t^{obj}-y_s]$ with a backward-warp process as:
\begin{equation}
    \left[x_{t}^{obj}, y_{t}^{obj}, 1\right]^{T} \sim \textbf{K}\left(\textbf{R}^{\dagger}-\frac{\textbf{t}^{\dagger} \textbf{n}^{T}}{\textbf{d}_{n}}\right) \textbf{K}^{-1}\left[x_{s}, y_{s}, 1\right]^{T},
\end{equation}
where $x_{s}$ and $y_{s}$ are uniformly sampled from a $H\times W$ grid. $\textbf{R}$ and $\textbf{t}$ are the rotation and translation from the source views $\bm{\mu}_{s}$ to the target view $\bm{\mu}_{t}^{obj}$, $\textbf{K}$ is the camera intrinsic, and $\textbf{n} = [0, 0, 1]$ is the normal vector. $\textbf{R}^{\dagger}$ and $\textbf{t}^{\dagger}$ are the inverses of $\textbf{R}$ and $\textbf{t}$, respectively. For background motion under view $\bm{\mu}_{t}^{bg}$, we compute the optical flow in the same way by $\textbf{F}_{n}^{bg}=[x_t^{bg}-x_s, y_t^{bg}-y_s]$. 

To separate dynamic objects from the background, we utilize an instance segmentation network $\Omega$ \cite{cheng2022masked} to extract the main object in the source image $\textbf{I}_s$ as:
\begin{equation}
    \textbf{M} = \Omega(\textbf{I}_s) \in \mathbb{R}^{H \times W},
\end{equation}
where $\textbf{M}$ is a binary mask to indicate the region of the object. The object mask $\textbf{M}_n$ on the $n$-th plane can be obtained using bilinear interpolation. Therefore, Equation \eqref{eq:eq1} for MPI construction can be re-formulated as:
\begin{align}
    & \{(\textbf{C}_n, \bm{\Sigma}_n, \textbf{D}_n, \textbf{M}_n, \textbf{F}_n^{obj}, \textbf{F}_n^{bg})\}^N_{n=1} \\
    & = \mathcal{F}(\textbf{I}_s, \textbf{D}_{s}) \notag \cup \mathcal{T}(\bm{\mu}_{s}, \bm{\mu}_{t}^{obj}, \bm{\mu}_{t}^{bg}) \notag \cup \mathcal{B}(\textbf{M}),
    \label{eq:eq3}
\end{align}
where $\mathcal{T}$ indicates the pixel shift operation to compute optical flow and $\mathcal{B}$ indicates bilinear interpolation. The constructed MPI thus can be viewed as a collection of planes with multiple properties, including colors, density, depth, flows, and object occupancy.

\subsection{Object-Independent Volume Rendering}

To render novel view images and the corresponding optical flow, volume rendering are used. We first formulate the volume rendering process. Assuming a virtual camera motion $\bm{\mu}_{t}$, each pixel ($x_t$, $y_t$) on the novel view image planes can be mapped to pixel ($x_s$, $y_s$) on $n$-th source MPI plane via homography function \cite{heyden2005multiple}:
\begin{equation}
    \left[x_{s}, y_{s}, 1\right]^{T} \sim \textbf{K}\left(\textbf{R}-\frac{\textbf{t} \textbf{n}^{T}}{\textbf{d}_{n}}\right) \textbf{K}^{-1}\left[x_{t}, y_{t}, 1\right]^{T},
    \label{eq:eq2}
\end{equation}
where $\textbf{R}$ and $\textbf{t}$ are the rotation and translation from the source view $\bm{\mu}_{s}$ to the target view $\bm{\mu}_{t}$, $\textbf{K}$ is the camera intrinsic, and $\textbf{n} = [0, 0, 1]$ is the normal vector. 

Then, we can render the novel view $\textbf{P}_t$ of a specific property under target view $\bm{\mu}_{t}$. Discrete intersection points between new planes and arbitrary rays passing through the scene and estimate integrals are used as:
\begin{equation}
    \textbf{P}_t=\sum_{n=1}^{N}\left(\textbf{P}_{n}^{t} \bm{\alpha}_{n}^{t} \prod_{m=1}^{n-1}\left(1-\bm{\alpha}_{m}^{t}\right)\right), 
\label{eq:render}
\end{equation}
where $\bm{\alpha}^{t}_n=\exp \left(-\bm{\delta}_{n} \bm{\sigma}^{t}_{n}\right)$ and $\bm{\delta}_{n}$ is the distance map between plane $n$ and $n+1$ and we set the initial depth of MPI planes uniformly spaced in disparity as in \cite{han2022single}. For example, the novel view image under target view $\bm{\mu}_{t}^{obj}$ can be rendered by re-formulating Equation \eqref{eq:render} as:
\begin{equation}
\textbf{I}_t^{obj}=\sum_{n=1}^{N}\left(\textbf{C}_{n}^{obj} \bm{\alpha}_{n}^{obj} \prod_{m=1}^{n-1}\left(1-\bm{\alpha}_{m}^{obj}\right)\right), 
\label{eq:image}
\end{equation}
where $\textbf{C}_{n}^{obj}$ is the $n^{th}$ color plane under target view $\bm{\mu}_{t}^{obj}$.

To make sure that the optical flow maps match the novel view image $\textbf{I}_t^{obj}$ perfectly, we propose to render optical flow $\textbf{F}_{s\rightarrow t}^{obj}$ in the same way as Equation \eqref{eq:render} by: 
\begin{equation}
    \textbf{F}_{s\rightarrow t}^{obj}=\sum_{n=1}^{N}\left(\textbf{F}_n^{obj} \bm{\alpha}_{n}^{obj} \prod_{m=1}^{n-1}\left(1-\bm{\alpha}_{m}^{obj}\right)\right), 
\label{eq:flow}
\end{equation}
where $\textbf{F}_{s\rightarrow t}^{obj}$ indicates the optical flow that matches novel view image $\textbf{I}_s^{obj}$. The pipeline implemented thus far models the optical flow of dynamic object motion. Then we propose applying separate virtual camera motions to the static background. We achieve this by rendering objects and backgrounds separately, following the rendering formulation described in Equation \eqref{eq:image}. We generate the novel view image of the background, $\textbf{I}_{t}^{bg}$. In this way, the frame $\textbf{I}_{t}^{obj}$ contains dynamic objects with 6DOF motion $\mu_{t}^{obj}$, while $\textbf{I}_{t}^{bg}$ contains static backgrounds influenced by camera motion $\mu_{t}^{bg}$. The final novel-view image $\textbf{I}_{t}$ is obtained by adding these two frames, $\textbf{I}_{t}^{obj}$ and $\textbf{I}_{t}^{bg}$, utilizing object masks. Similarly, the optical flow is obtained by adding the two optical flows, $\textbf{F}_{s\rightarrow t}^{obj}$ and $\textbf{F}_{s\rightarrow t}^{bg}$, which can be rendered in the same way with Equation \eqref{eq:flow}. Thend, we utilize the binary mask to indicate the region of the object. Thus the final optical flow $\textbf{F}_{s\rightarrow t}$ can be obtained by:
\begin{equation}
    \textbf{F}_{s\rightarrow t} = \textbf{M} \times \textbf{F}_{s\rightarrow t}^{obj} + (1 - \textbf{M}) \times \textbf{F}_{s\rightarrow t}^{bg},
\end{equation}
where $\textbf{F}_{s\rightarrow t}$ can serve as the ground truth labels for training optical flow estimation networks. And the novel view $\textbf{I}_{t}$ with dynamic objects can be obtained by:
\begin{equation}
    \textbf{I}_{t} = \textbf{M}_t^{obj} \times \textbf{I}^{obj}_{t} + (1 - \textbf{M}_t^{bg}) \times \textbf{I}^{bg}_{t},
    \label{eq:add-image}
\end{equation}
where $\textbf{M}_t^{obj}$ and $\textbf{M}_t^{bg}$ are the object masks generated using bilinear interpolation under target views $\mu_t^{obj}$ and $\mu_t^{bg}$. However, given the differences in viewing angles, simply adding two novel view images will result in unnatural occlusion due to separate motions between objects and background. In addition, areas that do not contain objects and backgrounds can also lead to empty holes. Therefore, we propose a Depth-Aware Inpainting module to solve this problem.

\begin{figure}[t]
\centering
\subfloat[Source Image]{\includegraphics[width=3.4in]{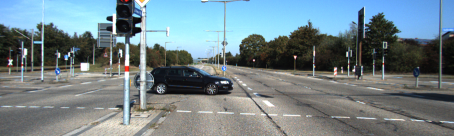}\label{11}}
\hfil
\subfloat[ + Camera Motion]{\includegraphics[width=3.4in]{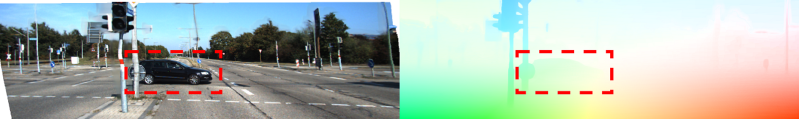}\label{22}}
\hfil
\subfloat[ + Object-Independent Volume Rendering]{\includegraphics[width=3.4in]{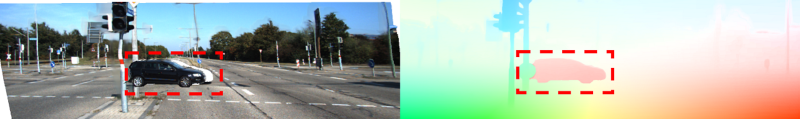}\label{33}}
\hfil
\subfloat[ + Depth-Aware Inpainting]{\includegraphics[width=3.4in]{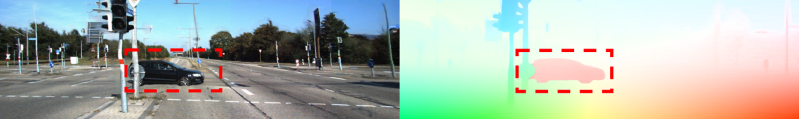}\label{44}}
\caption{Visualization of incrementally adding different modules to improve the realism of the generated data.}
\label{fig:incremental}
\end{figure}

\subsection{Depth-Aware Inpainting}

Although merged image through Equation \eqref{eq:add-image} gives a realistic visual effect, depth changes caused by different camera motion and object motion can also cause unnatural occlusions. To solve this problem, we use volume rendering to obtain the depth of the novel views, i.e., $\textbf{D}_{t}^{obj}$ and $\textbf{D}_{t}^{bg}$. We then utilize both depths to compute the occupation mask between the novel views:
\begin{equation}
    \textbf{M}_{occ}=(1-\textbf{M}_t^{bg}) \odot \textbf{M}_t^{obj} \odot (\textbf{D}_{t}^{bg}<\textbf{D}_{t}^{obj}),
\end{equation}
where $\textbf{M}_{occ}$ indicates the wrong background areas in front of the object. Therefore, we can restore the overlapped area between the object and the background in the new image $\textbf{I}_t$ by re-formulating Equation \eqref{eq:add-image}:
\begin{equation}
    \textbf{I}_{t} = \textbf{M}_t^{obj} \times \textbf{I}^{obj}_{t} + (1 - \textbf{M}_t^{bg} \odot \textbf{M}_{occ}) \times \textbf{I}^{bg}_{t},
    \label{eq:render3}
\end{equation}
which prevents wrong occlusions. After handling the occlusions, we use an advanced inpainting model to fill the holes that contain no object and no background. We first compute the mask of the area without object and background by:
\begin{equation}
    \textbf{M}_{inp}=(1-\textbf{M}_t^{bg}) \odot (1 - \textbf{M}_t^{obj}),
\end{equation}
which indicates the areas to fill. Then, a pre-trained Diffusion Inpainting model $\Pi$ \cite{Rombach_2022_CVPR} is utilized to fill the masked areas, and the final novel view image $\textbf{I}_{t}$ can be obtained by re-formulating Equation \eqref{eq:render3} as:
\begin{equation}
    \textbf{I}_{t} = \Pi(\textbf{M}_t^{obj} \times \textbf{I}^{obj}_{t} + (1 - \textbf{M}_t^{bg} \odot \textbf{M}_{occ}) \times \textbf{I}^{bg}_{t}, \textbf{M}_{inp}).
    \label{eq:render4}
\end{equation}

\begin{figure*}[!t]
\centering
{\includegraphics[height=2.1in]{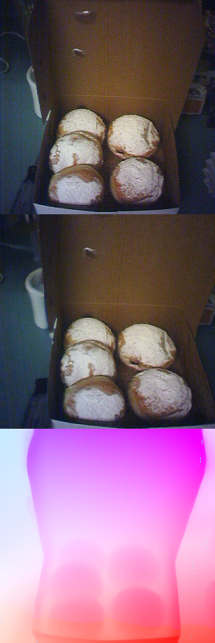}}
\hfil
{\includegraphics[height=2.1in]{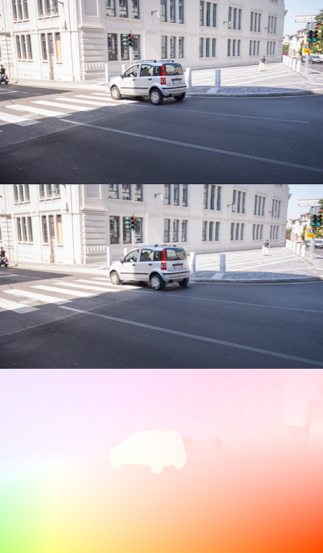}}
\hfil
{\includegraphics[height=2.1in]{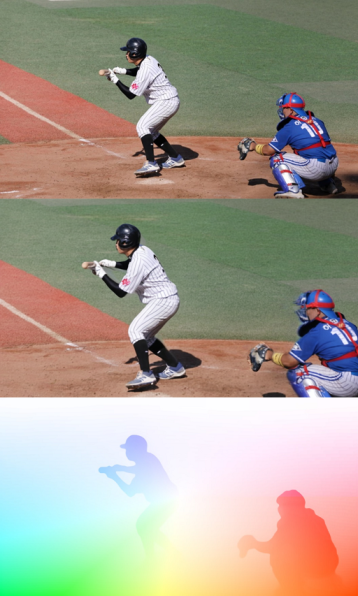}}
\hfil
{\includegraphics[height=2.1in]{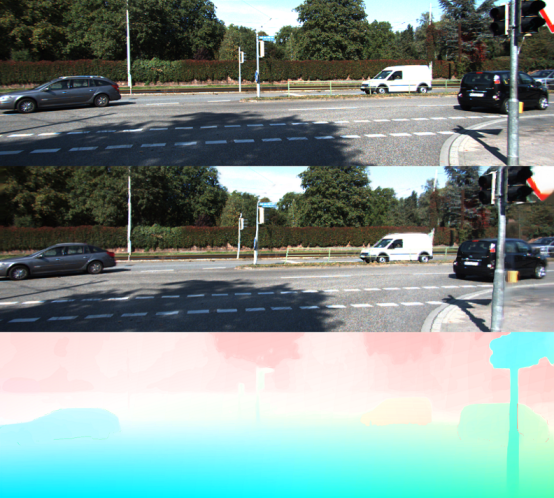}}
\hfil
{\includegraphics[height=2.1in]{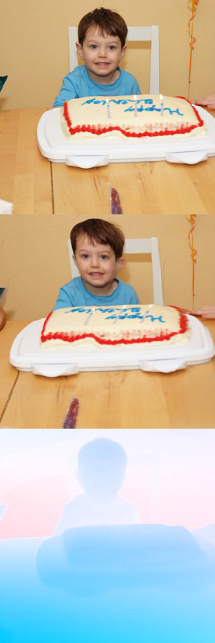}}
\hfil
{\includegraphics[height=2.1in]{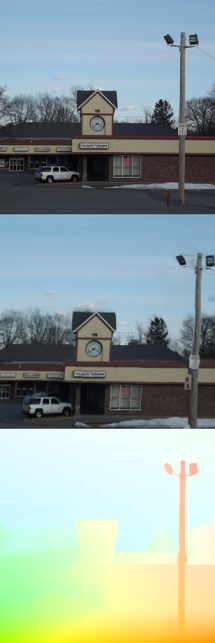}}
\hfil
\caption{Visualization of generated data using our proposed Flow-Anything from unlabeled single-view images for training. From top to bottom: source images, generated novel view images, and generated optical flow.}
\label{fig:example-data}
\end{figure*}

We provide a detailed illustration of the incremental effects of Flow-Anything, both with and without the Object-Independent Volume Rendering and Depth-Aware Inpainting modules, in subsection \ref{sub_ablation}. Novel view images and optical flow from single-view images can be generated using Flow-Anything with only camera motion, as depicted in Figures \ref{11} and \ref{22}. However, camera motion alone is insufficient to capture the complex optical flow present in real-world scenes. As demonstrated in Figure \ref{33}, the Object-Independent Volume Rendering module enhances motion realism by providing separate flows for dynamic objects and the background as in the real world. To further improve motion realism and address occlusions caused by dynamic object movement, we employ the Depth-Aware Inpainting module, as illustrated in Figure \ref{44}.

\subsection{Training Optical Flow Networks}

With generated data from our proposed optical flow data generation pipeline in Figure \ref{fig:framework-v2}. Optical flow estimation networks can be trained. To be specific, we basically follow the training pipeline of SEA-RAFT \cite{wang2024sea}. The networks take two images $\textbf{I}_s$ and $\textbf{I}_t$ as input and output estimated optical flow $\textbf{F}_{pred}$. Then, the generated optical flow $\textbf{F}_{s\rightarrow t}$ is used for supervised training. The training loss is:
\begin{equation}
    \mathcal{L}_{\text{prob}} = -\log p_{\theta}(\textbf{F}_{pred} = \textbf{F}_{s\rightarrow t} \mid \textbf{I}_s, \textbf{I}_t),
    \label{eq:loss1}
\end{equation}
where the probability density function $p_{\theta}$ is parameterized by the network. When training networks, we make no change to the setting of the training loss function. For example, when training SEA-RAFT, the loss is set to a mixture of two Laplace distributions as in \cite{wang2024sea}. 

\section{Experiments}

In this section, we first introduce the datasets and the details of our implementation. Then, we conduct various comparisons with the existing methods of optical flow data generation, supervised methods, and unsupervised methods, respectively. Finally, we show more discussions. We explain the nouns we use in the footnote, including \textbf{Flow-Anything} \footnote{Our proposed framework for optical flow training data generation, which only needs single-view images from the real world.}, \textbf{FA-Flow Dataset} \footnote{A large-scale optical flow training dataset using our Flow-Anything and mixed dataset resources as in Table \ref{tab:datasets}.}, and \textbf{FA-Flow (X)} \footnote{An optical flow dataset using our proposed Flow-Anything and only the images in dataset ``X" is used for data generation.}.

\subsection{Datasets}

In this subsection, we outline the datasets used for training and evaluation of our optical flow estimation networks. Models are pre-trained on FlyingChairs and FlyingThings3D by default. We also detail the mixed dataset we collected for data generation, in which the single-view images are utilized.

\noindent \textbf{FlyingChairs \cite{dosovitskiy2015flownet} and FlyingThings3D \cite{ilg2017flownet} (C+T)} are popular synthetic datasets used for training optical flow models. We denote these datasets as ``C" and ``T" respectively. ``C+T" indicates training on FlyingChairs first and then fine-tuning on FlyingThings3D. By default, we pre-train models on ``C+T" and then fine-tune it on the generated datasets.

\begin{table}[t] 
\centering
\caption{Labeled and unlabeled datasets used in our FA-Flow Dataset. We use our proposed Flow-Anything to generate the optical flow and novel view images using the single-view images from unlabeled datasets.}
\begin{tabular}{@{}cccc@{}}
\toprule
Real-World & Label & Dataset & \# Images \\ \midrule
& \multirow{5}{*}{Dense} & FlyingChairs \cite{dosovitskiy2015flownet} & 22,872 \\
&  & FlyingThings3D \cite{ilg2017flownet} & 21,818 \\
&  & MPI Sintel \cite{Butler2012} &  1,041 \\
&  & Virtual KITTI 2 \cite{cabon2020virtual} &  21,260 \\
&  & Spring \cite{Mehl2023_Spring} & 6,000 \\ \midrule
\multirow{2}{*}{\ding{51}} & \multirow{2}{*}{Sparse} & HD1K \cite{kondermann2015stereo} & 1,083 \\
&  & KITTI Flow 15 \cite{menze2015object} &  200 \\ \midrule
\multirow{9}{*}{\ding{51}} & \multirow{9}{*}{None} & BDD100k \cite{yu2020bdd100k} & 90,000 \\
&  & COCO \cite{lin2014microsoft} &  118,287 \\
&  & Google Landmarks \cite{weyand2020google} & 117,576 \\
&  & Nuscenes \cite{caesar2020nuscenes} &  93,475 \\
&  & SUN RGBD \cite{song2015sun} &  10,335 \\
&  & KITTI Raw \cite{Geiger2013IJRR} &  93,657 \\
&  & NYU V2 \cite{silberman2012indoor} &  45,205 \\
&  & Cityscapes \cite{cordts2016cityscapes} &  19,998 \\ 
&  & DAVIS \cite{perazzi2016benchmark} &  10,581 \\
\bottomrule
\end{tabular}
\label{tab:datasets}
\end{table}

\begin{table*}[ht]
\centering
\caption{Zero-shot cross-dataset validation results compared with data generation methods, using the same image sources. The results demonstrate the superior performance of our proposed Flow-Anything data generation method beyond other data generation methods. RAFT networks are trained on the generated data for fair comparison. In cases where RealFlow fails to work on single-view images from COCO, the table indicates ``N/A". The curly braces ``\{\}" represent the use of the unlabeled images from the evaluation set, which is the KITTI 15 training set in this table. MPI-Flow uses the same models as the competitors, and Flow-Anything uses more advanced monocular depth estimation \cite{depthanything} and inpainting models \cite{Rombach_2022_CVPR}.}

\begin{tabular}{@{}llcccccccc@{}}
\toprule
& ~ & \multicolumn{2}{c}{Sintel.C} & \multicolumn{2}{c}{Sintel.F} & \multicolumn{2}{c}{KITTI 12} & \multicolumn{2}{c}{KITTI 15} \\ \cmidrule(lr){3-4} \cmidrule(lr){5-6} \cmidrule(lr){7-8} \cmidrule(lr){9-10} 
\multirow{-2}{*}{Image Source} & \multirow{-2}{*}{Method} & EPE $\downarrow $ & $>$ 3 $\downarrow $ & EPE $\downarrow $ & $>$ 3 $\downarrow $ & EPE $\downarrow $ & Fl $\downarrow $ & EPE $\downarrow $ & Fl $\downarrow $ \\ \midrule
\multirow{4}{*}{COCO \cite{lin2014microsoft}} & Depthstillation \cite{aleotti2021learning} & 1.87 & 5.31 & 3.21 & 9.25 & 1.74 & 6.81 & 3.45 & 13.08 \\ 
& RealFlow \cite{han2022realflow} & N/A & N/A & N/A & N/A & N/A & N/A & N/A & N/A  \\ 
& MPI-Flow \cite{liang2023mpi} & 1.87 & 4.59 & 3.16 & 8.29 & 1.36 & 4.91 & 3.44 & 10.66  \\
& Flow-Anything & \textbf{1.81} & \textbf{4.47} & \textbf{3.14} & \textbf{8.22} & \textbf{1.29} & \textbf{4.90} & \textbf{3.38} & \textbf{10.32}  \\ \midrule
\multirow{4}{*}{DAVIS \cite{perazzi2016benchmark}} & Depthstillation \cite{aleotti2021learning} & 2.70 & 7.52 & 3.81 & 12.06 & 1.81 & 6.89 & 3.79 & 13.22  \\
& RealFlow \cite{han2022realflow} & 1.73 & 4.81 & 3.47 & 8.71 & 1.59 & 6.08 & 3.55 & 12.52  \\
& MPI-Flow \cite{liang2023mpi} & 1.79 & 4.77 & 3.06 & 8.56 & 1.41 & 5.36 & 3.32 & 10.47  \\
& Flow-Anything & \textbf{1.68} & \textbf{4.50} & \textbf{3.01} & \textbf{8.41} & \textbf{1.22} & \textbf{5.07} & \textbf{3.25} & \textbf{9.95}  \\ \midrule
\multirow{4}{*}{KITTI 15 Test \cite{menze2015object}} & Depthstillation \cite{aleotti2021learning} & 4.02 & 9.08 & 4.96 & 13.23 & 1.77 & 5.97 & 3.99  & 13.34 \\ 
& RealFlow \cite{han2022realflow} & 3.73 & 7.36 & 5.53 & 11.31 & 1.27 & 5.16 & 2.43 & 8.86  \\
& MPI-Flow \cite{liang2023mpi} & 2.25 & 5.25 & 3.65 & 8.89 & 1.24 & 4.51 & 2.16 & 7.30  \\
& Flow-Anything & \textbf{2.10} & \textbf{4.97} & \textbf{3.38} & \textbf{8.59} & \textbf{1.21} & \textbf{4.34} & \textbf{2.11} & \textbf{7.13}  \\ \midrule
\multirow{4}{*}{KITTI 15 Train \cite{menze2015object}} & Depthstillation \cite{aleotti2021learning} & 2.84 & 7.18 & 4.31 & 11.24 & 1.67 & 5.71 & \{2.99\} & \{9.94\} \\
& RealFlow \cite{han2022realflow} & 4.06 & 7.68 & 4.78 & 11.44 & 1.25 & 5.02 & \{2.17\} & \{8.64\}  \\
& MPI-Flow \cite{liang2023mpi} & 2.41 & 5.39 & 3.82 & 9.11 & 1.26 & 4.66 & \{1.88\} & \{7.16\}  \\ 
& Flow-Anything & \textbf{2.21} & \textbf{4.95} & \textbf{3.56} & \textbf{8.84} & \textbf{1.20} & \textbf{4.29} & \{\textbf{1.79}\} & \{\textbf{6.98}\}  \\ \bottomrule
\end{tabular}
\label{tab:dataset}
\end{table*}

\noindent \textbf{KITTI 2012 \cite{geiger2012we} and KITTI 2015 \cite{menze2015object} (K)} are renowned benchmarks for optical flow estimation, each with 200 labeled pairs for training and additional pairs for testing. We use ``K" to represent KITTI 2015. There are also multi-view extensions (4,000 for training and 3,989 for testing) datasets without ground truth, which We use to generate separate datasets for training and testing. By default, we evaluate the trained models on the training sets of KITTI 2012 and KITTI 2015, which serve as the validation sets, following previous work. Additionally, we present the fine-tuned results on the test sets of the KITTI 2015 official benchmark.

\noindent \textbf{Sintel \cite{butler2012naturalistic} (S)} is derived from the open-source 3D animated short film Sintel, denoted as ``S". The dataset includes 23 different scenes with RGB stereo images and disparity maps. Although it is not a real-world dataset, we use it to validate the model's generalization across different domains.

\noindent \textbf{Spring \cite{Mehl2023_Spring}} is a computer-generated, large, high-resolution, high-detail benchmark for scene flow, optical flow, and stereo. Based on rendered scenes from the open-source Blender movie ``Spring", it provides photo-realistic HD datasets with state-of-the-art visual effects and ground truth optical flow labels. We also use it to validate the model's generalization across different domains.

\noindent \textbf{Mixed datasets} are collected from a wide range of sources, as shown in Table \ref{tab:datasets}. FlyingChairs, FlyingThings3D, Sintel, Spring, and Virtual KITTI 2, despite having a significant number of images, do not represent real-world scenes. HD1K (``\textbf{H}") and KITTI Flow 15 are real-world datasets but provide only sparse labels and a limited amount of data, offering a more challenging and real-world dataset for network training. In contrast, the remaining datasets, from BDD100k to DAVIS, are real-world datasets without any labels. These datasets, which include a substantial number of images, are essential for cross-domain learning where our method learns to generate optical flow training without ground truth data. We select eight large-scale public datasets as our unlabeled sources for their diverse scenes. They contain more than 6M images in total. Figure \ref{fig:example-data} showcases some generated data from these unlabeled real-world datasets.

\subsection{Implementation Details}
Firstly, we describe the learning-based optical flow estimation models that are utilized in our experiments. Subsequently, we outline the experimental parameters and setup along with the evaluation formulation.

\noindent \textbf{Network Architectures.} To evaluate how effective our generated data are at training optical flow models, we select SEA-RAFT \cite{wang2024sea}, which represents state-of-the-art architecture for supervised optical flow estimation and has excellent generalization capability. We also show the results of another two representative network architectures, including RAFT and FlowFormer++. For monocular depth estimation and instance segmentation, we use Depth-Anything \cite{depthanything} and Mask2Former \cite{cheng2021mask2former}, which represent the state-of-the-arts. For the inpainting model, we use Stable Diffusion 2 Inpainting \cite{Rombach_2022_CVPR} for its superior performance on realism.

\noindent \textbf{Training Details.} By default, we train SEA-RAFT on generated data for 200K steps with a learning rate of $1\times10^{-4}$ and weight decay of $1\times10^{-5}$, batch size of $16$. Models. Image crops are set to $432\times960$ for training. For KITTI evaluation, the models are then fine-tuned to $288\times960$. For the rest of the setup, we use the official implementation of the SEA-RAFT without any modifications. 

\noindent \textbf{Virtual Camera Motion.} To generate the novel view images from single-view images, we adopt the same settings in \cite{aleotti2021learning} to build the virtual camera. For KITTI, Nuscenes, and Cityscapes, we empirically build the camera motion with three scalars where $t_x, t_y$ are in $[-0.2, 0.2]$ and $t_z$ are in $[0.1, 0.35]$. This is more in line with the camera parameters of the autonomous driving scenario. We build the camera rotation with three Euler angles $a_x, a_y, a_z$ in $[-\frac{\pi}{90}, \frac{\pi}{90}]$.

\begin{figure}[t]
\centering
\subfloat[Source Image from KITTI 15]{\includegraphics[width=3.4in]{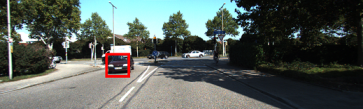}\label{sub2-baseline}}
\hfil
\subfloat[Depthstill.]{\includegraphics[width=0.8in]{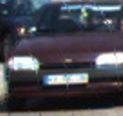}\label{sub2-ds}}
\hfil
\subfloat[RealFlow]{\includegraphics[width=0.8in]{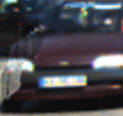}\label{sub2-baseline2}}
\hfil
\subfloat[MPI-Flow]{\includegraphics[width=0.8in]{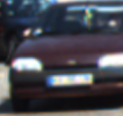}\label{sub2-ours}}
\hfil
\subfloat[Ours]{\includegraphics[width=0.8in]{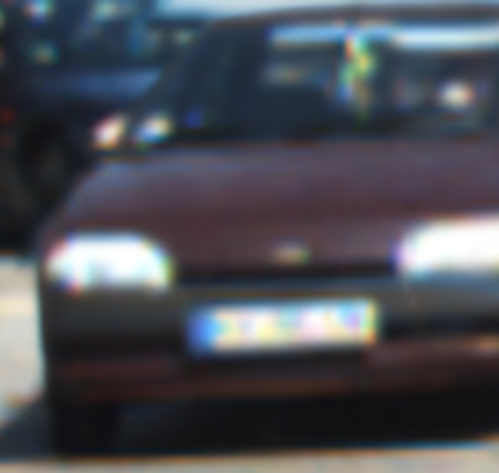}\label{sub2-fa}}
\caption{Visualization of generated images using Depthstillation, RealFlow, MPI-Flow, and our Flow-Anything from KITTI 15.}
\label{fig:samples-real}
\end{figure}

\begin{table*}[ht]
\centering
\caption{Comprehensive comparison of optical flow estimation accuracy of models with both supervised and unsupervised methods. The results demonstrate the superiority of our generated large-scale optical flow dataset, namely FA-Flow Dataset, beyond synthetic datasets and unsupervised learning methods. Our models are trained on the optical flow generated from the mixed dataset using our proposed Flow-Anything. ``-" indicates no results reported. For Sintel and KITTI, models are pre-trained on ``C+T". For the Spring dataset, models are pre-trained on ``C+T+S+K+H". $^{\dag}$ indicates that datasets for validation are not included in our mixed dataset when training. It is important to note that unsupervised methods generally require continuous frames for training and fail in single-view images, which constrains their applicability to larger datasets. In comparison, our method requires only single-view images for data generation, allowing for the use of a broader range of image resources.}

\begin{tabular}{@{}cllccccccccc@{}}
\toprule
& \multirow{2}{*}{Training Dataset} & \multirow{2}{*}{Method} & \multicolumn{2}{c}{Sintel} & \multicolumn{2}{c}{KITTI 15} & \multicolumn{2}{c}{KITTI 12} & \multicolumn{2}{c}{Spring} \\
\cmidrule(lr){4-5} \cmidrule(lr){6-7} \cmidrule(lr){8-9} \cmidrule(lr){10-11}
& & & Clean $\downarrow$ & Final $\downarrow$ & EPE $\downarrow$ & Fl-all $\downarrow$ & EPE $\downarrow$ & Fl-all $\downarrow$ & 1px $\downarrow$ & EPE $\downarrow$ \\ \midrule
\multirow{16}*{\rotatebox{90}{Supervised Methods}} & \multirow{13}{*}{C+T (+K+S+H)} & PWC-Net \cite{sun2018pwc} & 2.55 & 3.93 & 10.4 & 33.7 & 4.14 & 21.3 & - & - \\
& & RAFT \cite{teed2020raft}  & 1.43 & 2.71 & 5.04 & 17.4 & 2.15 & 9.30 & 4.78 & 0.448 \\
& & GMA \cite{jiang2021learning} & 1.30 & 2.74 & 4.69 & 17.1 & 1.99 & 9.27 & 4.76 & 0.443 \\
& & SKFlow \cite{sun2022skflow} & 1.22 & 2.46 & 4.27 & 15.5 & 1.89 & 8.35 & 4.52 & 0.408 \\
& & FlowFormer \cite{huang2022flowformer} & 1.01 & 2.40 & 4.09 & 14.7 & 1.80 & 7.20 & 4.50 & 0.470 \\
& & DIP \cite{zheng2022dip} & 1.30 & 2.82 & 4.29 & 13.7 & - & - & 4.27 & 0.463 \\
& & EMD-L \cite{deng2023explicit} & 0.88 & 2.55 & 4.12 & 13.5 & - & - & - & - \\
& & CRAFT \cite{sui2022craft} & 1.27 & 2.79 & 4.88 & 17.5 & 2.01 & 9.33 & 4.80 & 0.448 \\
& & RPKNet \cite{morimitsu2024recurrent} & 1.12 & 2.45 & 3.79 & 13.0 & 1.68 & 6.83 & 4.47 & 0.416 \\
& & GMFlowNet \cite{zhao2022global} & 1.14 & 2.71 & 4.24 & 15.4 & 1.89 & 8.43 & 4.29 & 0.433 \\
& & SEA-RAFT \cite{wang2024sea} & 1.21 & 4.04 & 4.29 & 14.2 & 1.96 & 7.26 & 4.06 & 0.406 \\
& & GMFlow \cite{xu2022gmflow} & 1.08 & 2.48 & 7.77 & 23.4 & 3.46 & 14.9 & 29.4 & 0.930 \\
& & GMFlow+ \cite{xu2023unifying} & 0.91 & 2.74 & 5.74 & 17.6 & 2.47 & 10.0 & 4.29 & 0.433 \\  \cmidrule(lr){2-11}
& YouTube-VOS \cite{xu2018youtube} + C+T (+K+S+H) & FlowFormer++ \cite{shi2023flowformer++} & 0.90 & 2.30 & 3.93 & 14.1 & 1.78 & 7.35 & 4.48 & 0.447 \\ \cmidrule(lr){2-11} 
& Tartan \cite{wang2020tartanair} + C+T (+K+S+H) & SEA-RAFT \cite{wang2024sea} & 1.19 & 4.11 & 3.62 & 12.9 & 2.01 & 7.51 & 3.88 & 0.406 \\ \midrule
\multirow{13}*{\rotatebox{90}{Unsupervised Methods}} & \multirow{13}{*}{C+T+Unlabeled Datasets} & UnFlow-CSS \cite{meister2018unflow}  & - & 7.91 & 8.10 & - & 3.29 & - & - & - \\
& & DDFlow \cite{liu2019ddflow} & 2.92 & 3.98 & 5.72 & 14.2 & 2.35 & - & - & - \\
& & SelFlow \cite{liu2019selflow} & 2.88 & 3.87 & 4.84 & 14.1 & 1.69 & - & - & - \\
& & SimFlow \cite{im2020unsupervised} & 2.96 & 3.57 & 5.19 & 13.3 & - & - & - & - \\
& & UFlow \cite{jonschkowski2020matters} & 3.01 & 4.09 & 2.84 & 9.39 & 1.68 & - & - & - \\
& & UPFlow \cite{luo2021upflow} & 2.33 & 2.67 & 2.45 & - & 1.27 & - & - & - \\
& & ARFlow \cite{liu2020learning} & 2.73 & 3.69 & 2.85 & - & 1.44 & - & - & - \\
& & SemARFlow \cite{yuan2023semarflow} & - & - & 2.18 & - & 1.28 & - & - & - \\
& & SemiFlow \cite{im2022semi} & 1.30 & 2.46 & 3.35 & 11.1 & - & - & - & - \\
& & UnSAMFlow \cite{yuan2024unsamflow} & 2.21 & 3.07 & 2.01 & - & 1.26 & - & - & - \\
& & SS-AutoFlow \cite{huang2022self} & 1.51 & 2.30 & 1.96 & 6.26 & - & - & - & - \\
& & SMURF \cite{stone2021smurf} & 1.71 & 2.58 & 2.00 & 6.42 & - & - & - & - \\
\midrule
\rowcolor{graycolor} \cellcolor{white} &  \cellcolor{white} & RAFT & 1.28 & 2.43 & 1.76 & 6.43 & 1.15 & 4.33 & 4.50 & 0.428 \\
\rowcolor{graycolor} \cellcolor{white} & \cellcolor{white} & FlowFormer++ & \textbf{0.86} & \textbf{2.20} & 1.89 & 6.92 & 1.06 & 4.02 & 4.43 & 0.431 \\
\rowcolor{graycolor} \cellcolor{white} \multirow{-3}{*}{\rotatebox{90}{Ours}} & \cellcolor{white} \multirow{-3}{*}{FA-Flow Dataset$^{\dag}$} & SEA-RAFT & 1.06 & 2.22 & \textbf{1.64} & \textbf{5.27} & \textbf{0.98} & \textbf{3.09} & \textbf{3.79} & \textbf{0.398} \\
\bottomrule
\end{tabular}
\label{tab:zero-shot}
\end{table*}

\subsection{Comparison with Data Generation Methods}

In this subsection, we validate the performance of our proposed Flow-Anything data generation framework on public benchmarks, comparing it with other representative data generation methods, including Depthstillation \cite{aleotti2021learning}, RealFlow \cite{han2022realflow}, and MPI-Flow \cite{liang2023mpi}. To evaluate Flow-Anything, we build training sets from four different datasets: COCO, DAVIS, KITTI 15 multi-view training set, and KITTI 15 multi-view test set. For fair comparisons, we conduct experiments with the same setup as our competitors and do not use extra data from our mixed dataset. Specifically, for DAVIS and KITTI, we generate four data pairs for each image to match RealFlow, which trains RAFT for four EM iterations with four times the amount of data. For COCO, we follow the setup used in Depthstillation. Since Depthstillation does not provide details for the KITTI 15 train, we use its default settings to get the results. Trained models are evaluated on the training sets of Sintel, KITTI 12, and KITTI 15. We report the evaluation results of the best-performing models for Depthstillation and RealFlow. Additionally, we conduct cross-dataset experiments where RAFT is trained on one generated dataset and evaluated on another to measure zero-shot performance.

As shown in Table \ref{tab:dataset}, with the same amount of generated data, our Flow-Anything achieves significant improvements and better generalization across multiple datasets. When trained and tested with the same KITTI 15 Train image source for data generation, our method outperforms the second-best method by a remarkable margin. When trained and tested with different image sources, Flow-Anything performs better than competitors in all evaluation settings. 

\noindent \textbf{Qualitative Results.} Figure \ref{fig:samples-real} showcases the generated images from various methods utilizing real-world images, as detailed in Table \ref{tab:dataset}. For this comparison, we use images from the KITTI 15 dataset as the source image input. The qualitative analysis focuses on the visual fidelity and realism of the generated images. The images generated by RealFlow \cite{han2022realflow} and Depthstillation \cite{aleotti2021learning} display artifacts that degrade the realism of the images. In contrast, MPI-Flow and Flow-Anything produce images that are more realistic than those generated by the other two methods. The qualitative results demonstrate that Flow-Anything significantly outperforms Depthstillation, RealFlow, and MPI-Flow in terms of generating realistic images. In addition, Flow-Anything reduces unnatural artifacts compared to MPI-Flow. This may benefit from more accurate depth estimates, as well as a more natural inpainting effect driven by the diffusion model. 

\begin{table}[t]
\centering
\caption{Comparison of models trained on our FA-Flow Dataset with synthetic datasets. For synthetic datasets, we choose Virtual KITTI 2 (VK2) dataset, which is specially designed and aligned with KITTI 15 dataset. The results indicate that our FA-Flow Dataset with diverse real-world images can defeat well-designed synthetic data.}
\begin{tabular}{@{}llccccc@{}}
\toprule
\multirow{2}{*}{Training Dataset} & \multirow{2}{*}{Method} & \multicolumn{2}{c}{KITTI 15}  \\
\cmidrule(lr){3-4}
& & EPE $\downarrow$ & Fl-all $\downarrow$ \\
\midrule
\multirow{14}{*}{C+T+VK2 Dataset \cite{cabon2020virtual}} & PWC-Net \cite{sun2018pwc} & 7.14 & 19.5 \\
& RAFT \cite{teed2020raft}  & 2.45 & 7.90 \\
& GMA \cite{jiang2021learning} & 2.33 & 7.62 \\
& SKFlow \cite{sun2022skflow} & 2.28 & 7.30 \\
& FlowFormer \cite{huang2022flowformer} & 2.41 & 8.17 \\
& FlowFormer++ \cite{shi2023flowformer++} & 2.34 & 8.44 \\
& DIP \cite{zheng2022dip} & 2.27 & 8.01 \\
& EMD-L \cite{deng2023explicit} & 2.56 & 9.24 \\
& CRAFT \cite{sui2022craft} & 2.55 & 9.06 \\
& RPKNet \cite{morimitsu2024recurrent} & 2.35 & 6.70 \\
& GMFlowNet \cite{zhao2022global} & 2.19 & 7.40 \\
& GMFlow \cite{xu2022gmflow} & 2.85 & 10.7 \\
& GMFlow+ \cite{xu2023unifying} & 2.25 & 7.20 \\
& SEA-RAFT \cite{wang2024sea} & 2.27 & 6.72 \\
\midrule
\rowcolor{graycolor} FA-Flow Dataset & SEA-RAFT & \textbf{1.64} & \textbf{5.27} \\
\bottomrule
\end{tabular}
\label{tab:finetune-vk}
\end{table}

\subsection{Comprehensive Comparison}
\label{subsec:modules}

In this subsection, we comprehensively compare the optical flow estimation accuracy of models trained on our large-scale generated data, namely FA-Flow Dataset, with other representative datasets and methods. For comprehensive comparison, we compare the state-of-the-arts, including supervised, semi-supervised, and unsupervised methods. Given that our proposed flow-Anything achieves the best performance in the data generation comparison, we use our method to synthesize optical Flow data from single-view images in the mixed datasets to train the models. We first validate the performance on two representative datasets in the real world: KITTI 15 and KITTI 12. We basically follow the experiment setup of SEA-RAFT \cite{wang2024sea}. We do not include the dataset for training when using it for evaluation. As shown in Table \ref{tab:zero-shot}, our Flow-Anything outperforms the best models across extensive scenes. For example, when tested on the real-world datasets KITTI 15 and KITTI 12, models trained on our FA-Flow Dataset gains remarkable improvements on three representative model architectures, including RAFT, FlowFormer++, and SEA-RAFT. For example, when train on our FA-Flow Dataset, SEA-RAFT gains remarkable improvements $-1.98$ on EPE and $-7.63$ on Fl-all on the KITTI 15 dataset.

We also show the generalization of our model on the unseen animated images and conduct experiments on another two representative synthetic datasets: Sintel and Spring. Our model also shows the best performance compared with the state-of-the-art supervised and unsupervised methods. Given that the SEA-RAFT architecture trained on our FA-Flow dataset achieves the best performance in most cases, especially on the real-world dataset, we use it as our default network. These results underscore the robustness and versatility of models trained on our FA-Flow dataset across various domains and datasets. 

\begin{table}[t]
\centering
\caption{Comparison of models pre-trained on our FA-Flow Dataset with synthetic datasets, when fine-tuned on the real-world dataset, i.e., KITTI 15 training set. Models are validated on the KITTI 15 test sets using the official benchmark. By default, all supervised methods have undergone ``C+T+S+K+H" pre-training process and then fine-tuned on KITTI 15 training set. ``-" indicates no results on the leader-board and not reported in the paper. The results indicate that our FA-Flow Dataset can serve as a powerful pre-training dataset, allowing the models to show better performance when fine-tuned on in-domain training datasets with ground truth labels.}
\begin{tabular}{@{}llccc@{}}
\toprule
\multirow{2}{*}{Training Dataset} & \multirow{2}{*}{Method} & \multicolumn{3}{c}{KITTI 15 (Test)} \\ \cmidrule(lr){3-5} 
 & & Fl-all $\downarrow$ & Fl-bg $\downarrow$ & Fl-fg $\downarrow$ \\
\midrule
 \multirow{14}{*}{C+T+S+K+H} & PWC-Net \cite{sun2018pwc} & 7.72 & 7.69 & 7.88 \\
 & RAFT \cite{teed2020raft} & 5.10 & 4.74 & 6.87 \\
 & GMA \cite{jiang2021learning} & 5.15 & - & - \\
 & SKFlow \cite{sun2022skflow} & 4.85 & 4.55 & 6.39 \\
 & FlowFormer \cite{huang2022flowformer} & 4.68 & 4.37 & 6.18 \\
 & DIP \cite{zheng2022dip} & 4.21 & 3.86 & 5.96 \\
 & EMD-L \cite{deng2023explicit} & 4.49 & 4.16 & 6.15 \\
 & CRAFT \cite{sui2022craft} & 4.79 & 4.58 & 5.85 \\
 & RPKNet \cite{morimitsu2024recurrent} & 4.64 & 4.63 & 4.69 \\
 & GMFlowNet \cite{zhao2022global} & 4.79 & 4.39 & 6.84 \\
 & GMFlow \cite{xu2022gmflow} & 9.32 & 9.67 & 7.57 \\
 & GMFlow+ \cite{xu2023unifying} & 4.49 & 4.27 & 5.60 \\
 & Flowformer++ \cite{shi2023flowformer++} & 4.52 & - & - \\
 & SEA-RAFT \cite{wang2024sea} & 4.64 & 4.47 & 5.49 \\
 \midrule
 \rowcolor{graycolor} FA-Flow Dataset & SEA-RAFT & \textbf{3.66} & \textbf{3.46} & \textbf{4.67} \\ 
\bottomrule
\end{tabular}
\label{tab:finetune-main}
\end{table}

\begin{table}[t]
\centering
\caption{Comparison with unsupervised optical flow estimation methods on the KITTI 15 test sets when trained on unlabeled images from KITTI 15 training set. Models do not use the optical flow labels from the KITTI 15 training sets for training or fine-tuning. ``-" indicates no results on the leader board and not reported in the paper. The results indicate that models trained on our generated datasets show better robustness when using only unlabeled images from the target domain.}
\begin{tabular}{@{}llccc@{}}
\toprule
\multirow{2}{*}{Training Dataset} & \multirow{2}{*}{Method} & \multicolumn{3}{c}{KITTI 15 (Test)} \\ \cmidrule(lr){3-5} 
&  & Fl-all $\downarrow$ & Fl-bg $\downarrow$ & Fl-fg $\downarrow$ \\
\midrule
\multirow{12}{*}{Unlabeled KITTI 15} & UnFlow \cite{meister2018unflow} & 11.11 & 10.15 & 15.93 \\ 
& DDFlow \cite{liu2019ddflow} & 14.29 & 13.08 & 20.40 \\ 
& SelFlow \cite{liu2019selflow} & 14.19 & 12.68 & 21.74 \\ 
& SimFlow \cite{im2020unsupervised} & 13.38 & 12.60 & 17.27 \\ 
& UFlow \cite{jonschkowski2020matters} & 11.13 & 9.78 & 17.87 \\ 
& UPFlow \cite{luo2021upflow} & 9.38 & - & - \\ 
& ARFlow \cite{liu2020learning} & 11.80 & - & - \\ 
& SemARFlow \cite{yuan2023semarflow} & 8.38 & 7.48 & 12.91 \\
& UnSAMFlow \cite{yuan2024unsamflow} & 7.83 & 6.40 & 14.98 \\ 
& SS-AutoFlow \cite{huang2022self} & 6.76 & 5.90 & 11.09 \\ 
& SMURF \cite{stone2021smurf} & 6.83 & 6.04 & 10.75 \\	\midrule
\rowcolor{graycolor} FA-Flow (KITTI 15) & SEA-RAFT & \underline{6.11} & \underline{5.47} & \underline{9.09} \\
\rowcolor{graycolor} FA-Flow Dataset & SEA-RAFT & \textbf{4.91} & \textbf{4.40} & \textbf{7.48} \\	
\bottomrule
\end{tabular}
\label{tab:finetune-main-un}
\end{table}

\begin{figure*}[t]
\centering
\subfloat[Frames]{\includegraphics[width=1.42in]{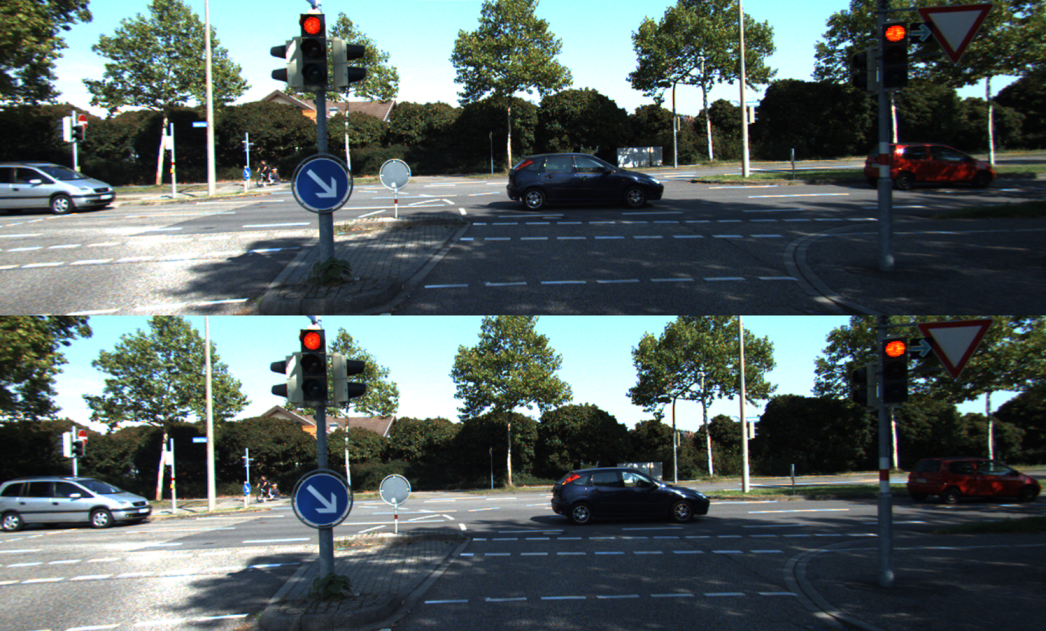}}
\hfil
\subfloat[UnFlow \cite{meister2018unflow}]{\includegraphics[width=1.42in]{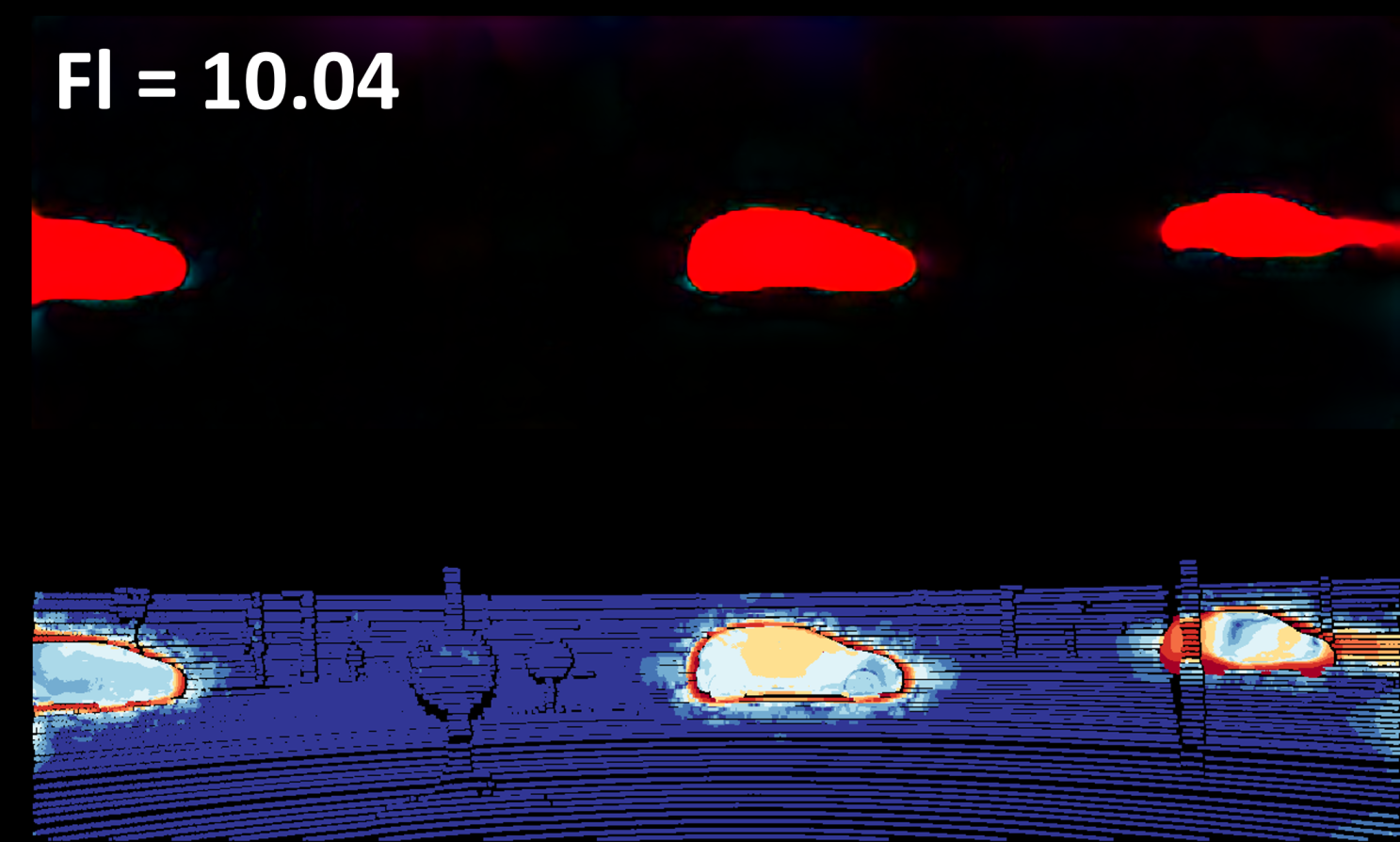}}
\hfil
\subfloat[DDFlow \cite{liu2019ddflow}]{\includegraphics[width=1.42in]{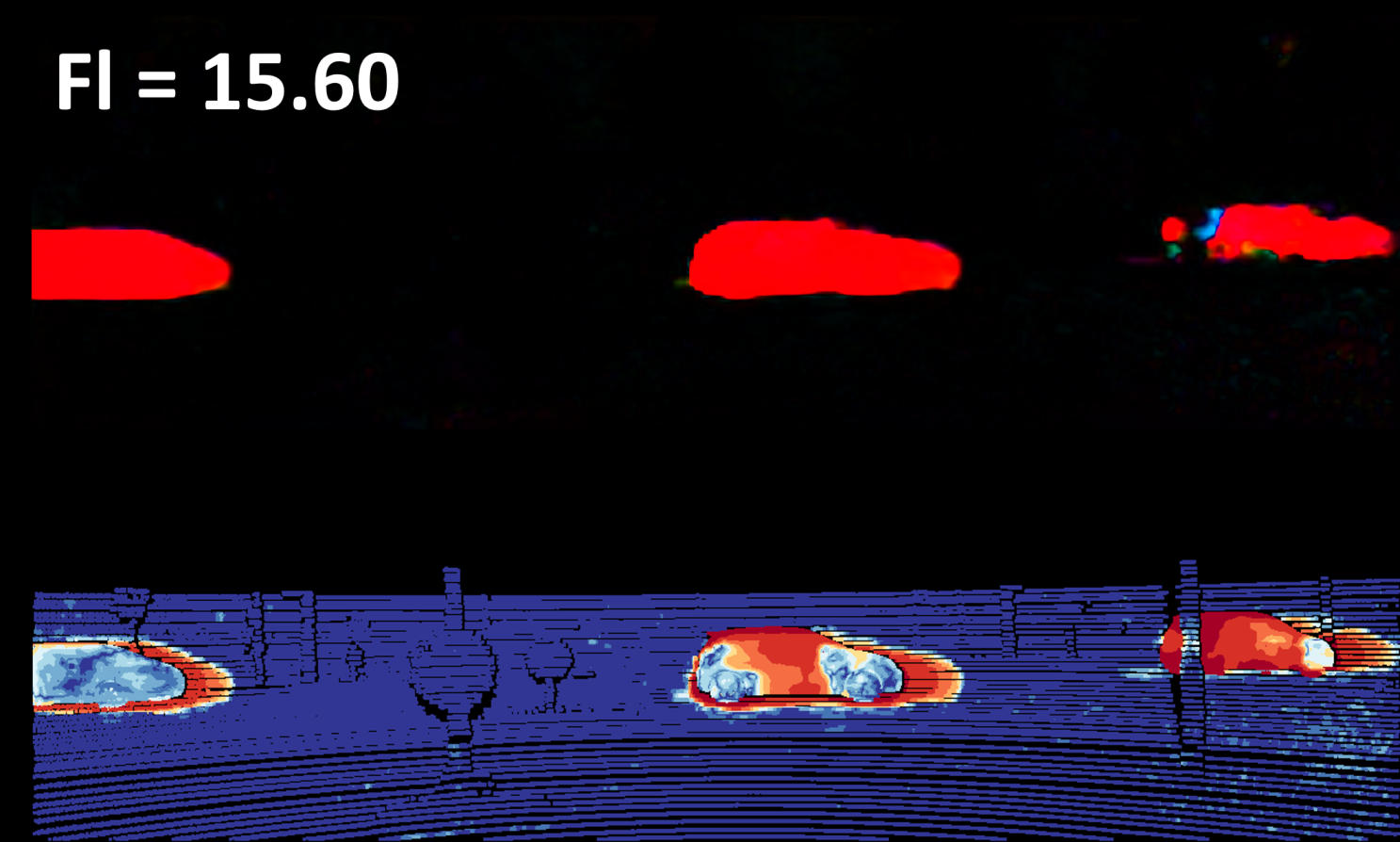}}
\hfil
\subfloat[SelFlow \cite{liu2019selflow}]{\includegraphics[width=1.42in]{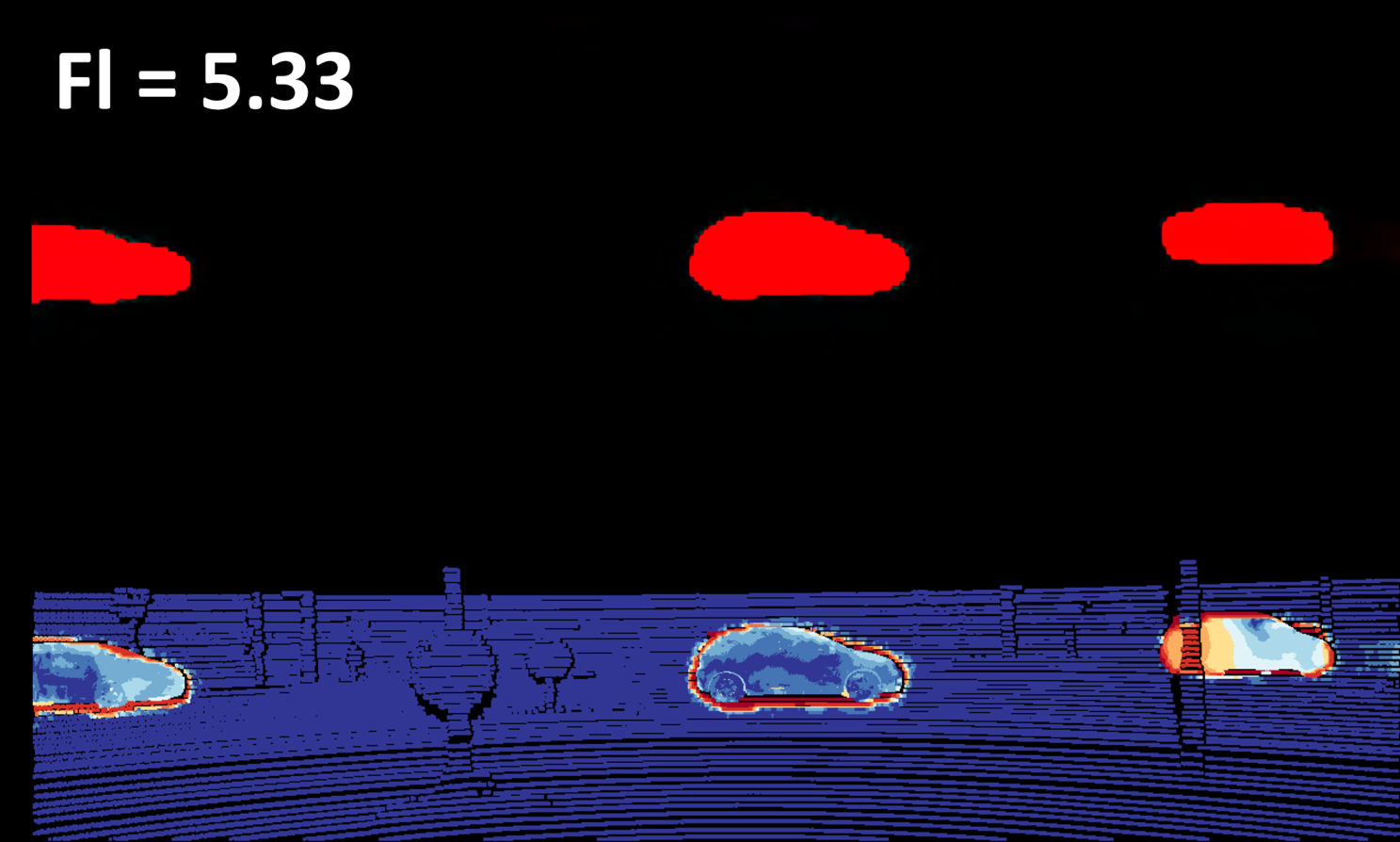}}
\hfil
\subfloat[SimFlow \cite{im2020unsupervised}]{\includegraphics[width=1.42in]{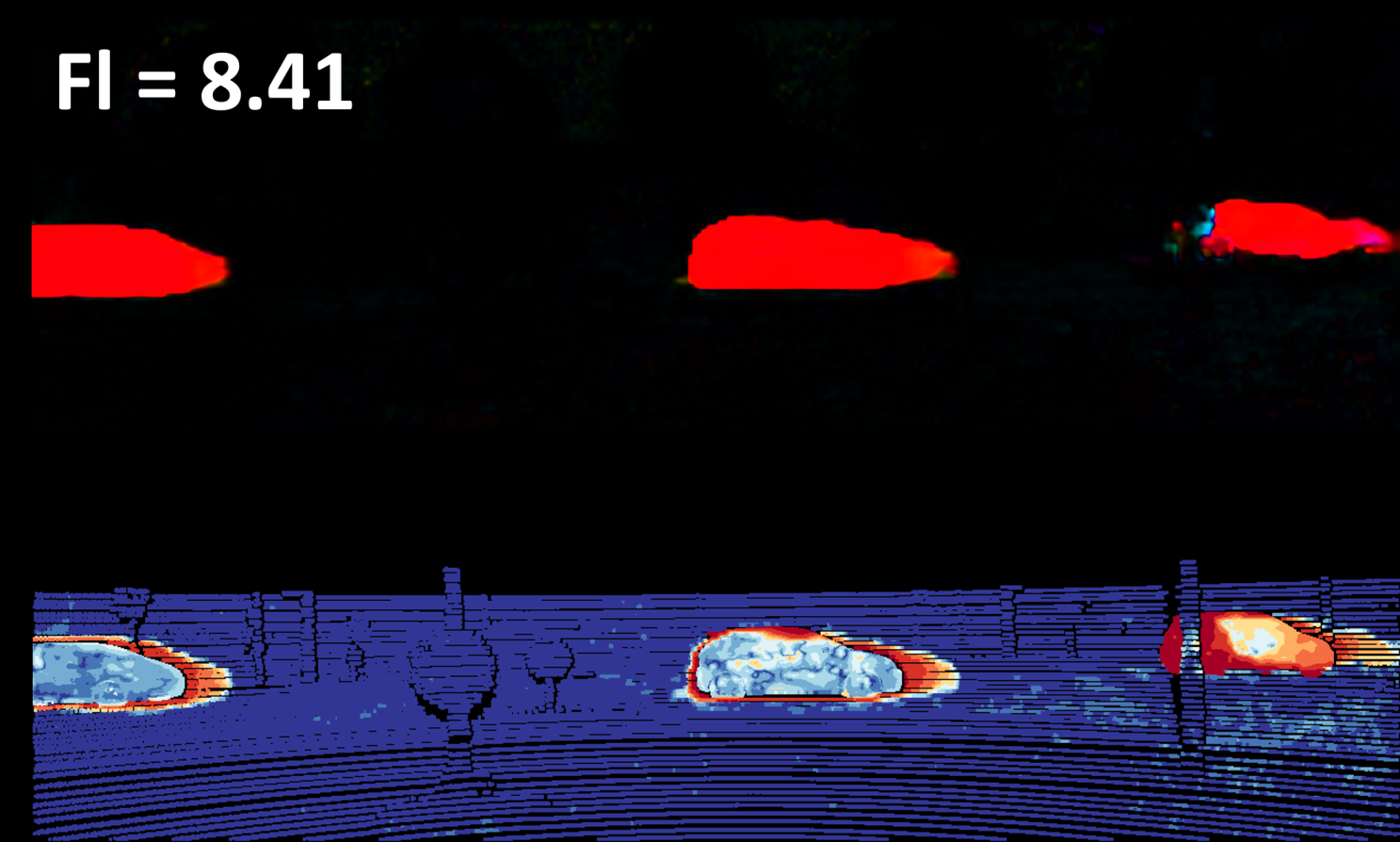}}
\hfil
\subfloat[UFlow \cite{jonschkowski2020matters}]{\includegraphics[width=1.42in]{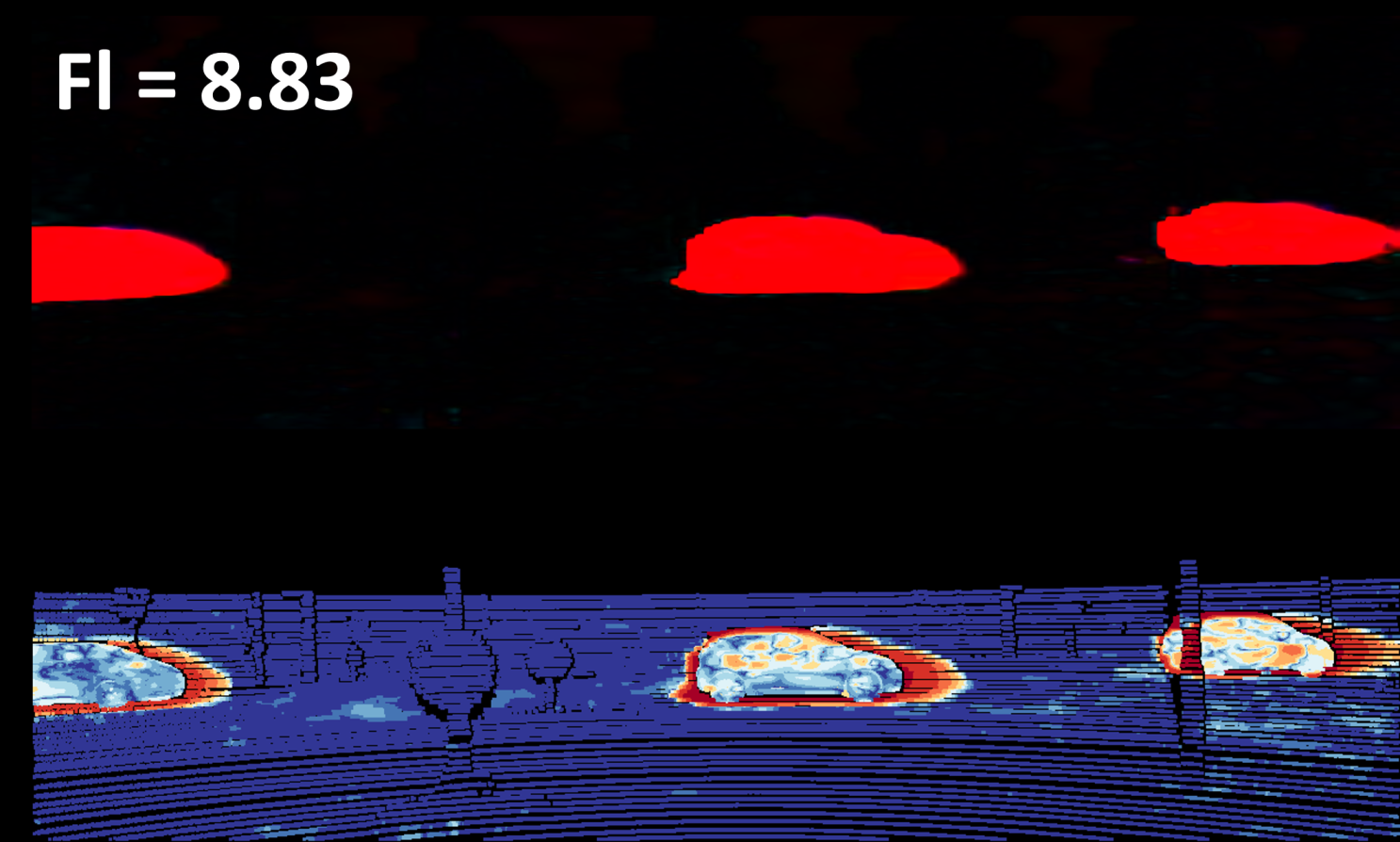}}
\hfil
\subfloat[SemARFlow \cite{yuan2023semarflow}]{\includegraphics[width=1.42in]{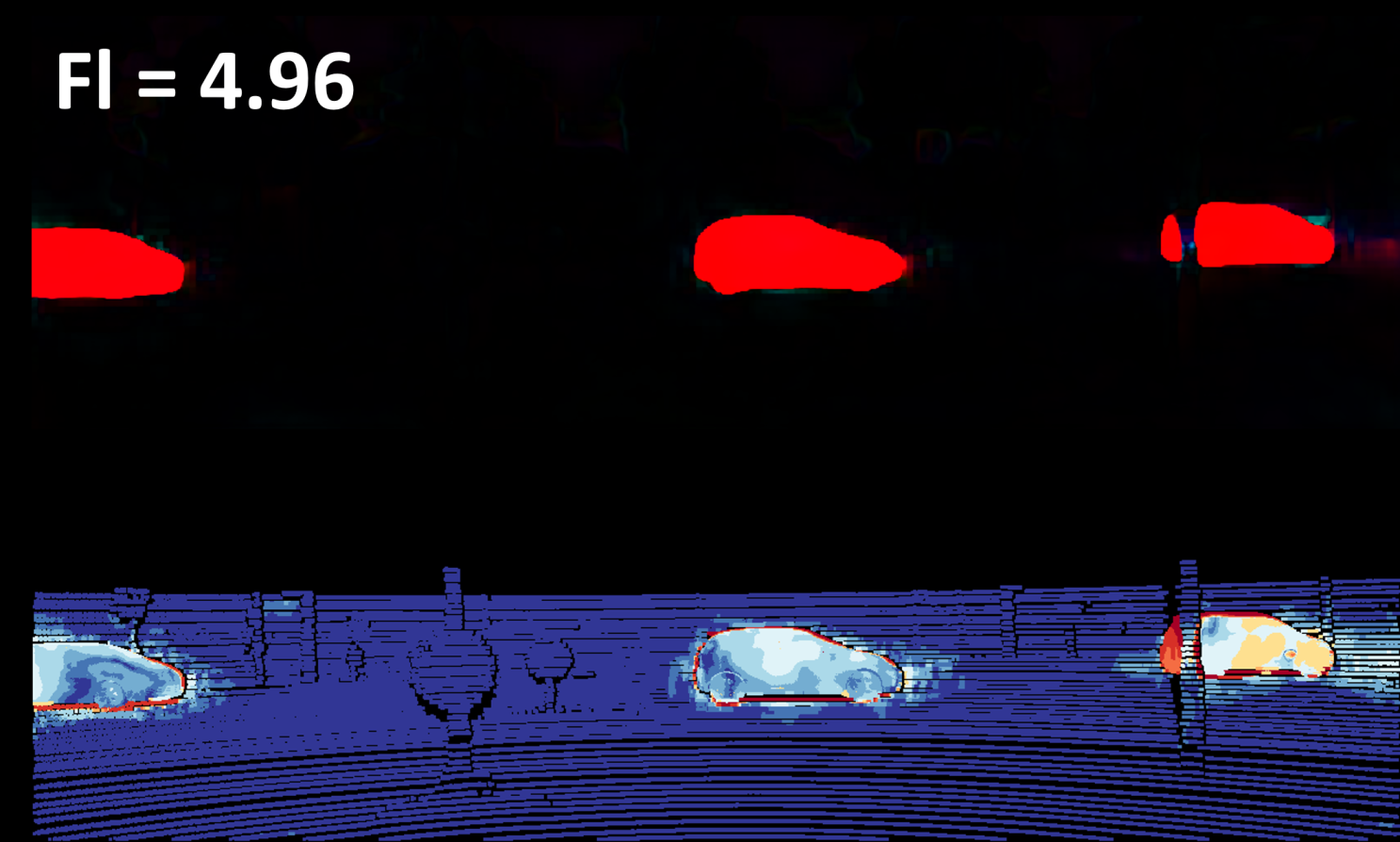}}
\hfil
\subfloat[UnSAMFlow \cite{yuan2024unsamflow}]{\includegraphics[width=1.42in]{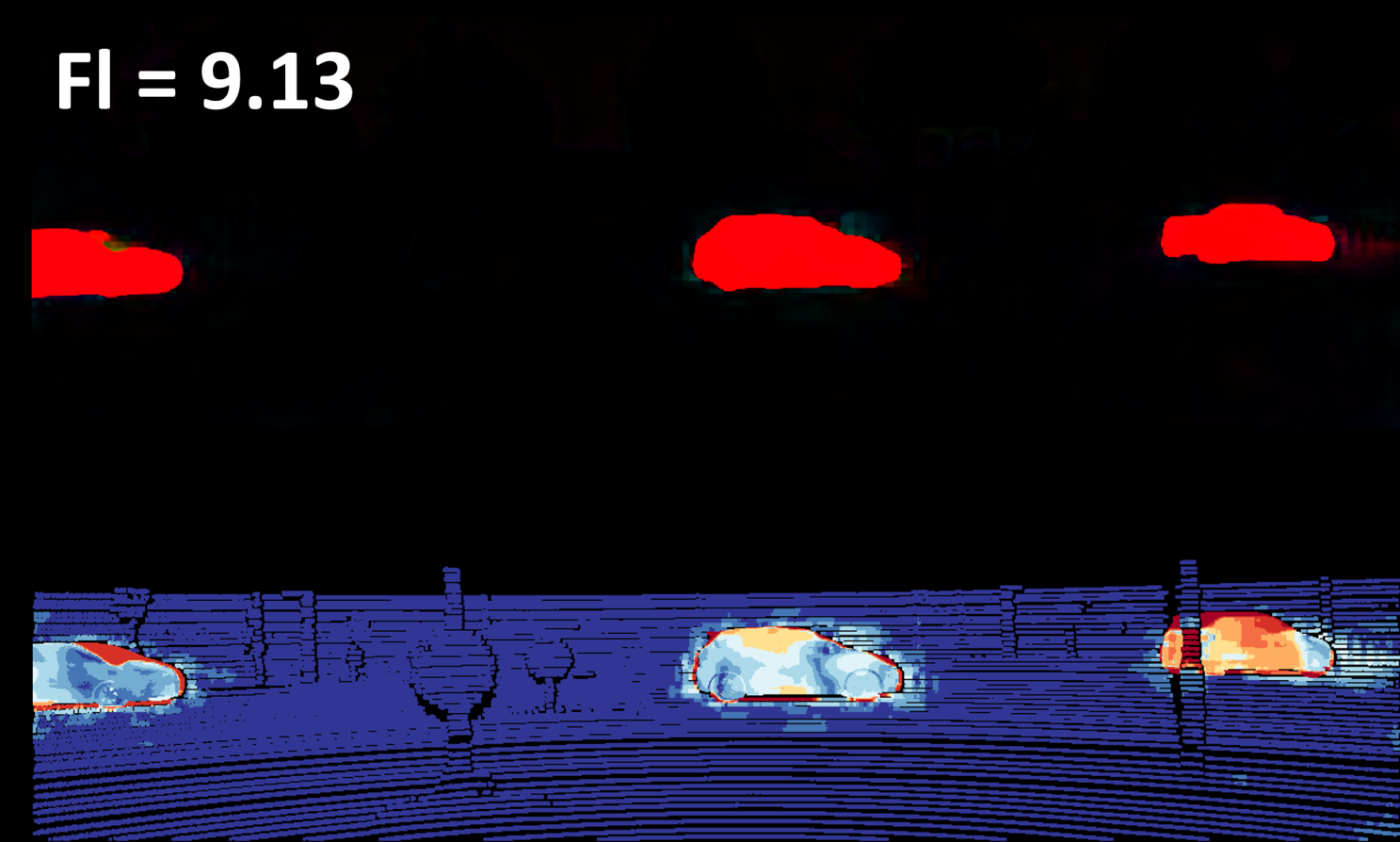}}
\hfil
\subfloat[SMURF \cite{stone2021smurf}]{\includegraphics[width=1.42in]{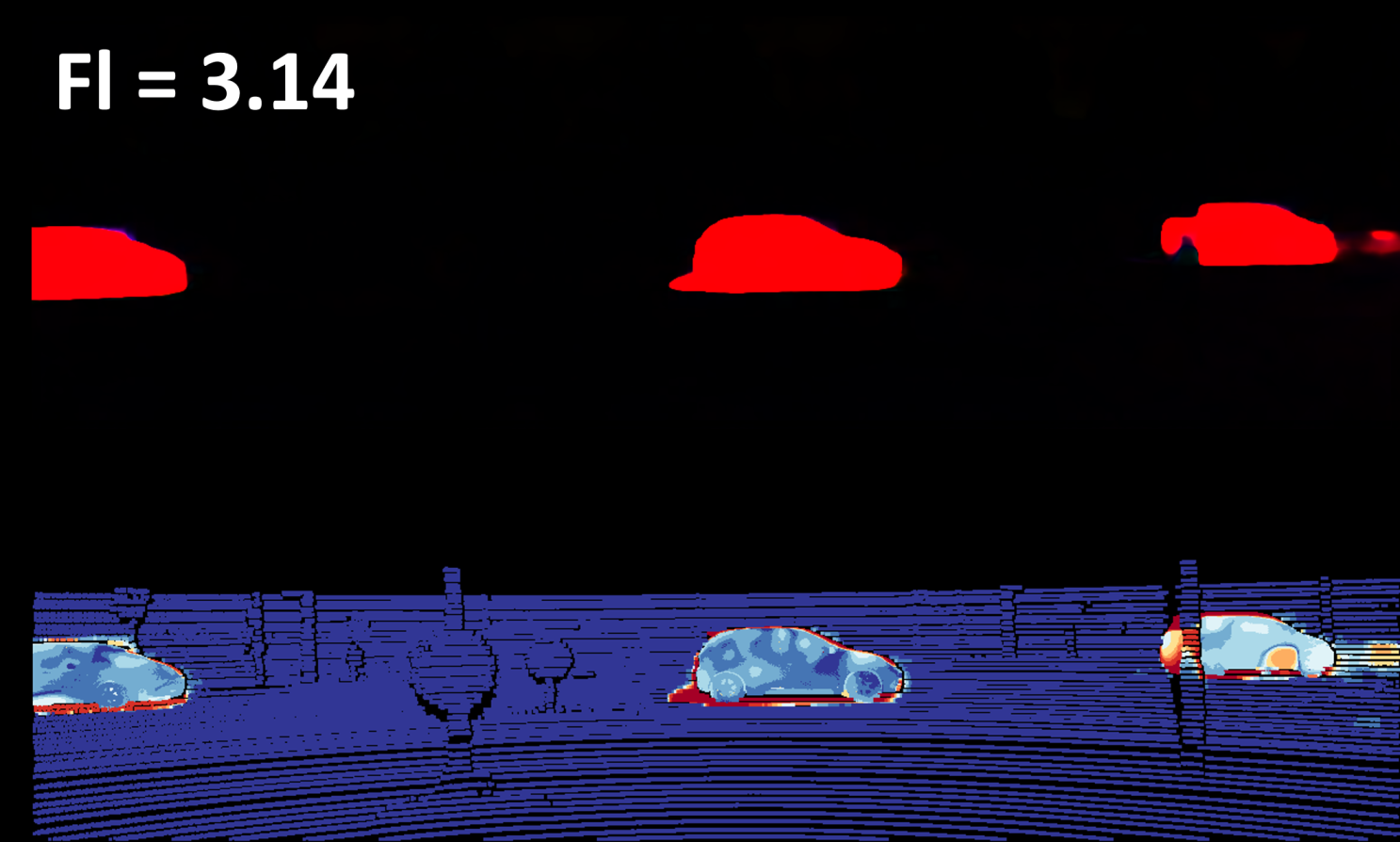}}
\hfil
\subfloat[Ours]{\includegraphics[width=1.42in]{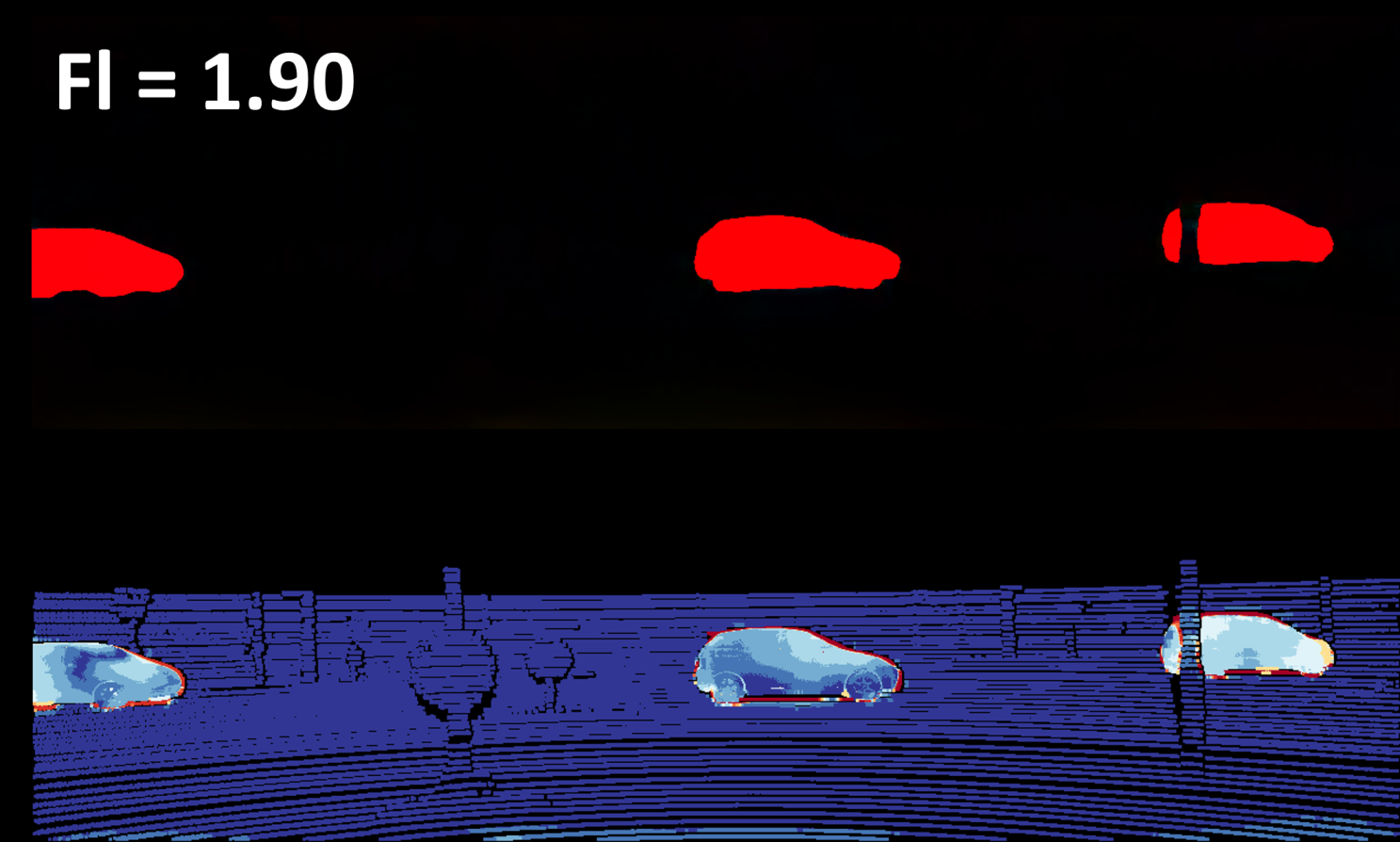}}
\caption{Unsupervised qualitative results on the KITTI-2015 test. All methods are trained without real labels on KITTI. We use the examples of the methods that are public on the KITTI Vision Benchmark. From top to bottom: predicted optical flow and error maps (sample frame \#000009).}
\label{fig:example-optical-flow}
\end{figure*}

\begin{figure*}[t]
\centering
\subfloat[Frames]{\includegraphics[width=1.42in]{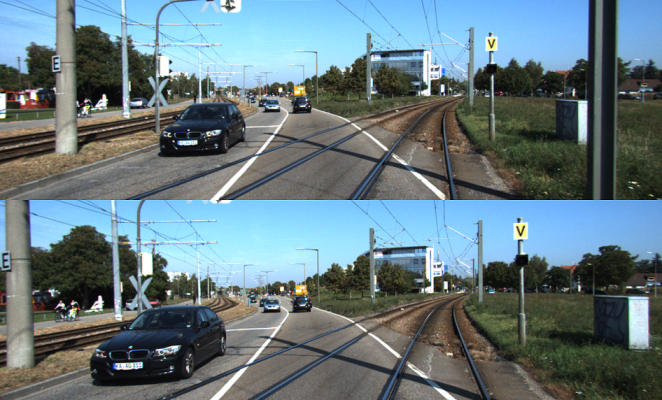}}
\hfil
\subfloat[UnFlow \cite{meister2018unflow}]{\includegraphics[width=1.42in]{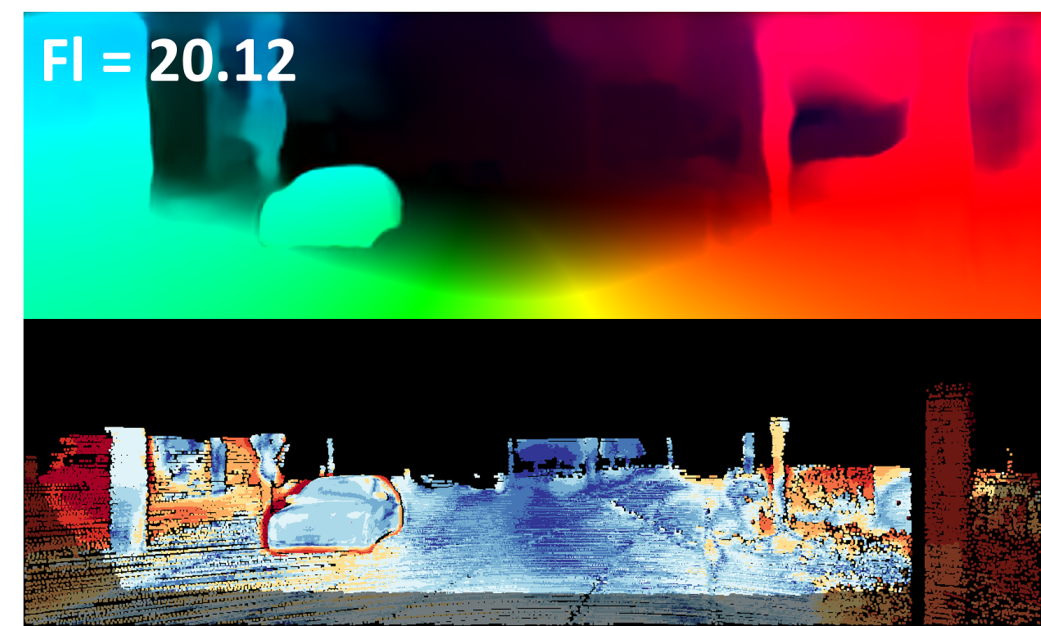}}
\hfil
\subfloat[DDFlow \cite{liu2019ddflow}]{\includegraphics[width=1.42in]{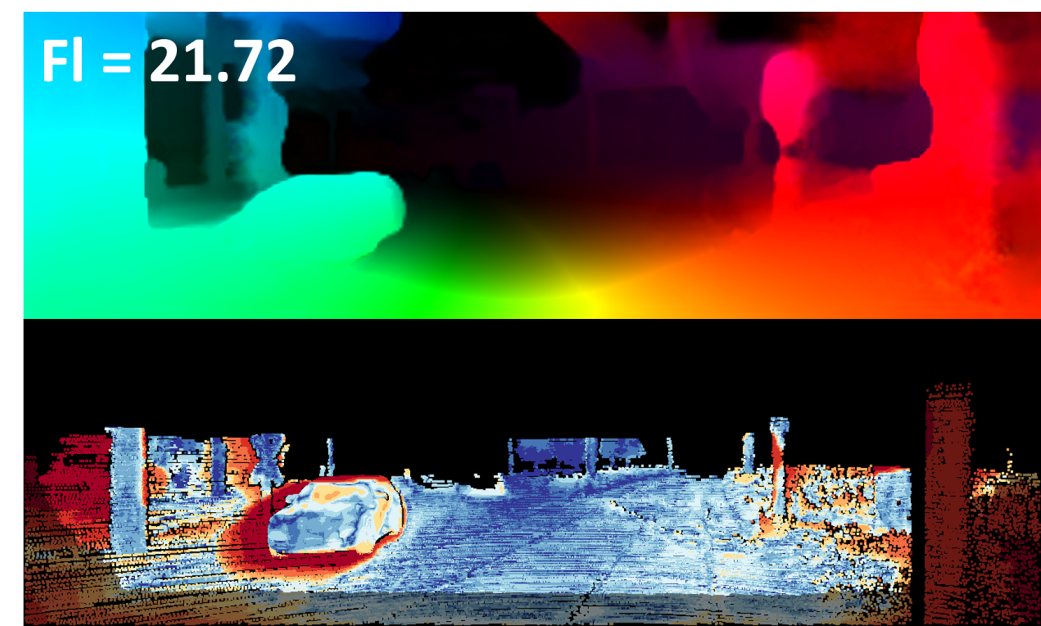}}
\hfil
\subfloat[SelFlow \cite{liu2019selflow}]{\includegraphics[width=1.42in]{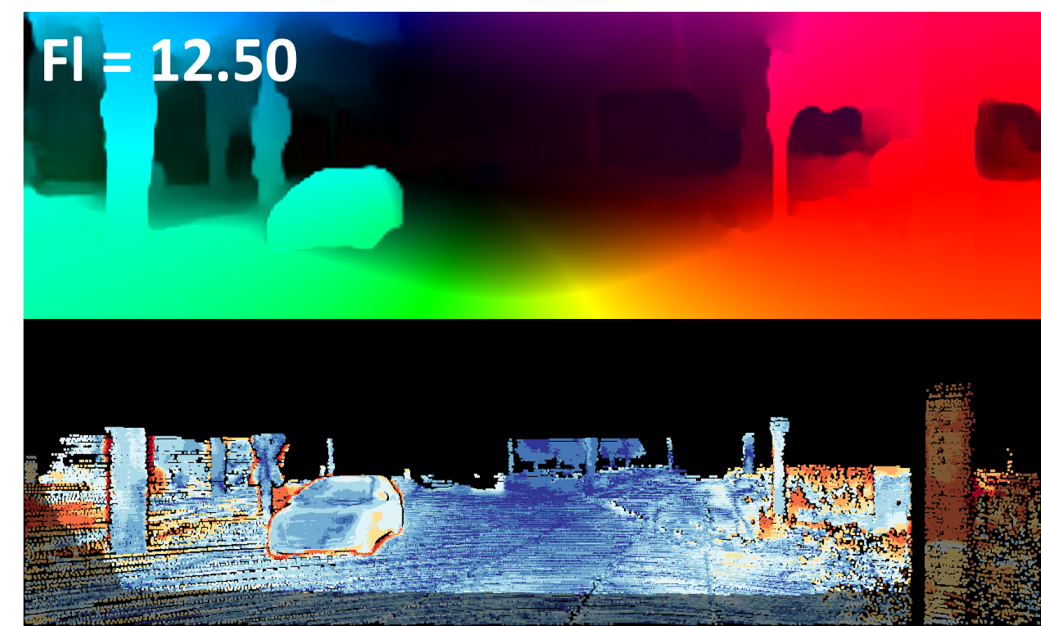}}
\hfil
\subfloat[SimFlow \cite{im2020unsupervised}]{\includegraphics[width=1.42in]{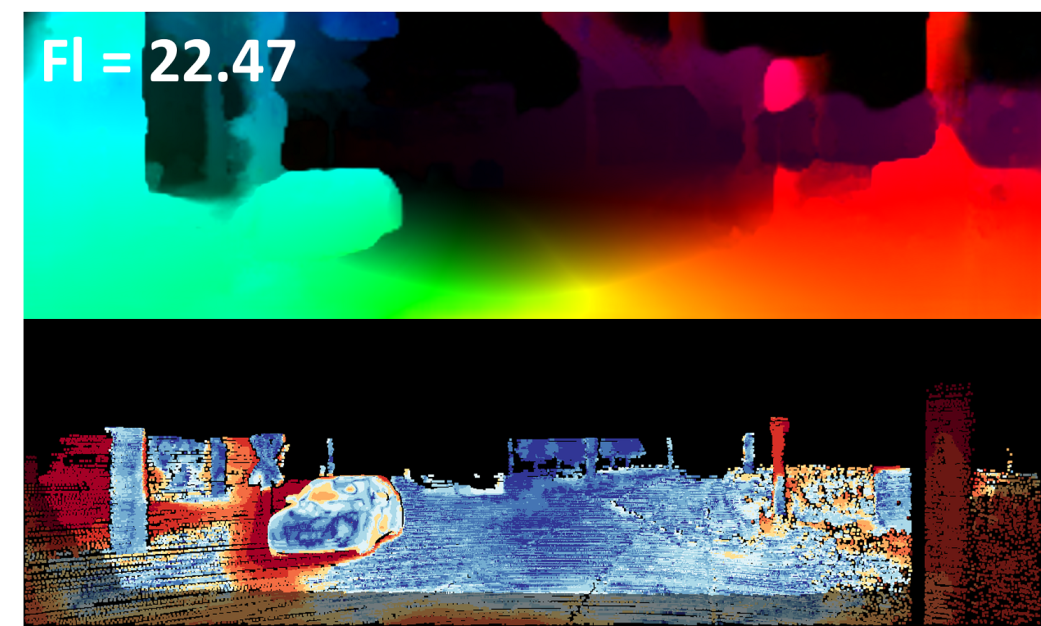}}
\hfil
\subfloat[UFlow \cite{luo2021upflow}]{\includegraphics[width=1.42in]{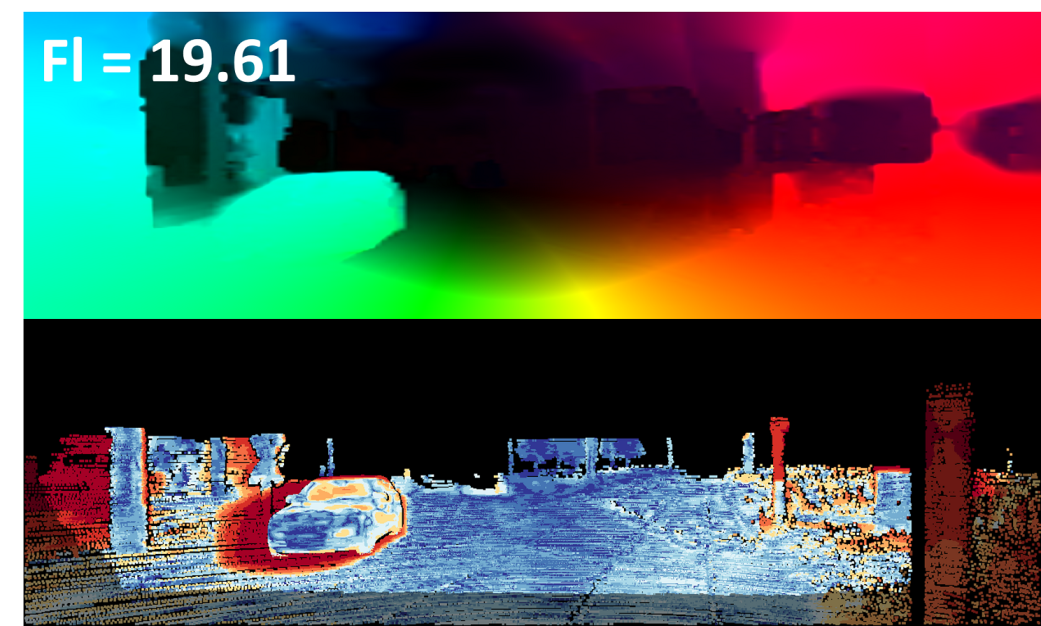}}
\hfil
\subfloat[SemARFlow \cite{yuan2023semarflow}]{\includegraphics[width=1.42in]{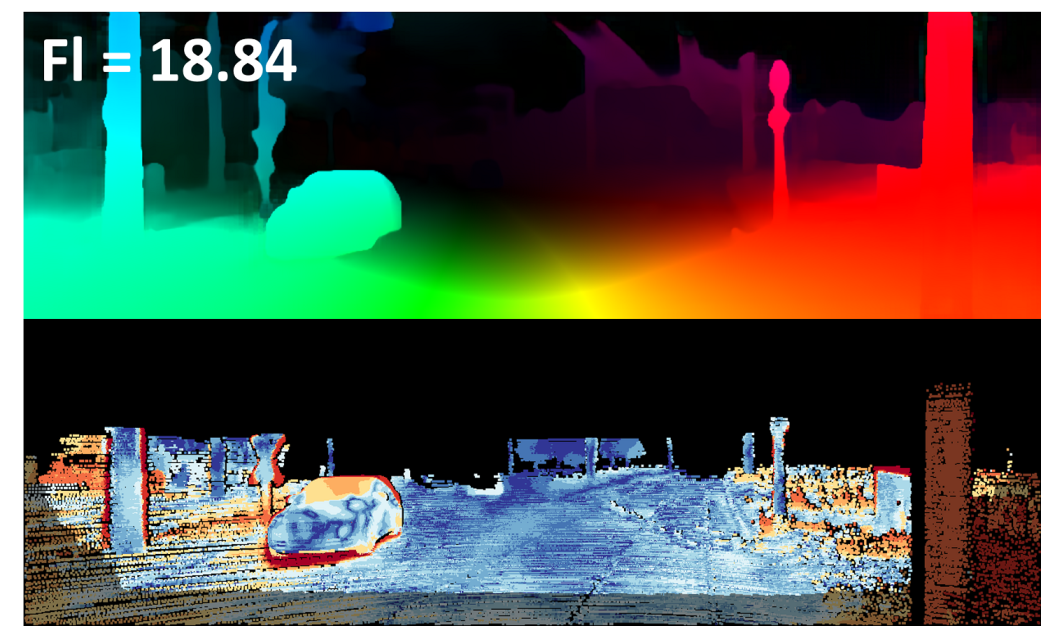}}
\hfil
\subfloat[UnSAMFlow \cite{yuan2024unsamflow}]{\includegraphics[width=1.42in]{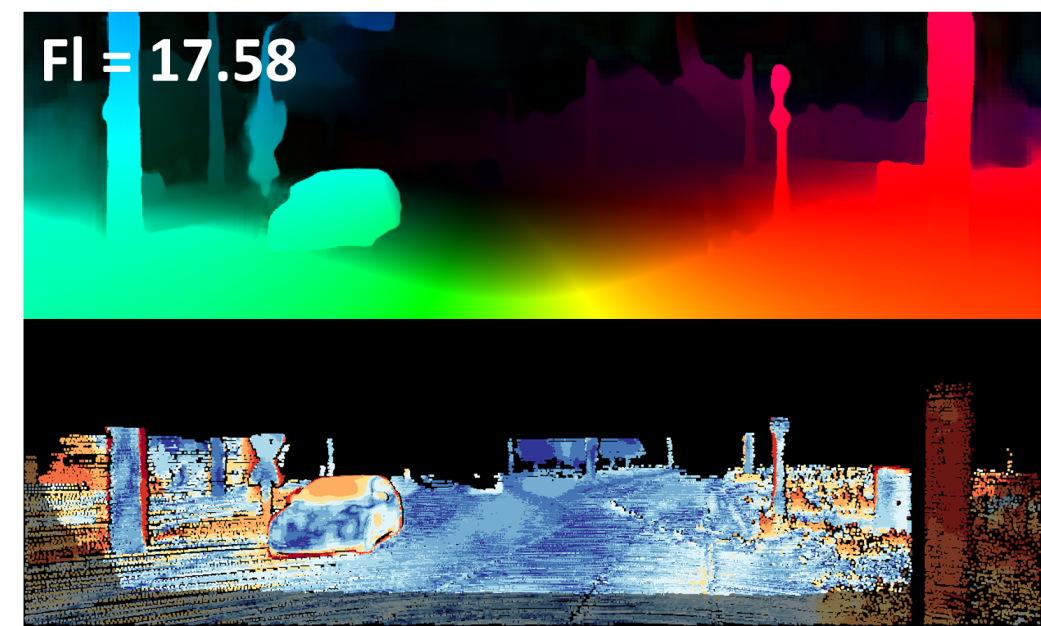}}
\hfil
\subfloat[SMURF \cite{stone2021smurf}]{\includegraphics[width=1.42in]{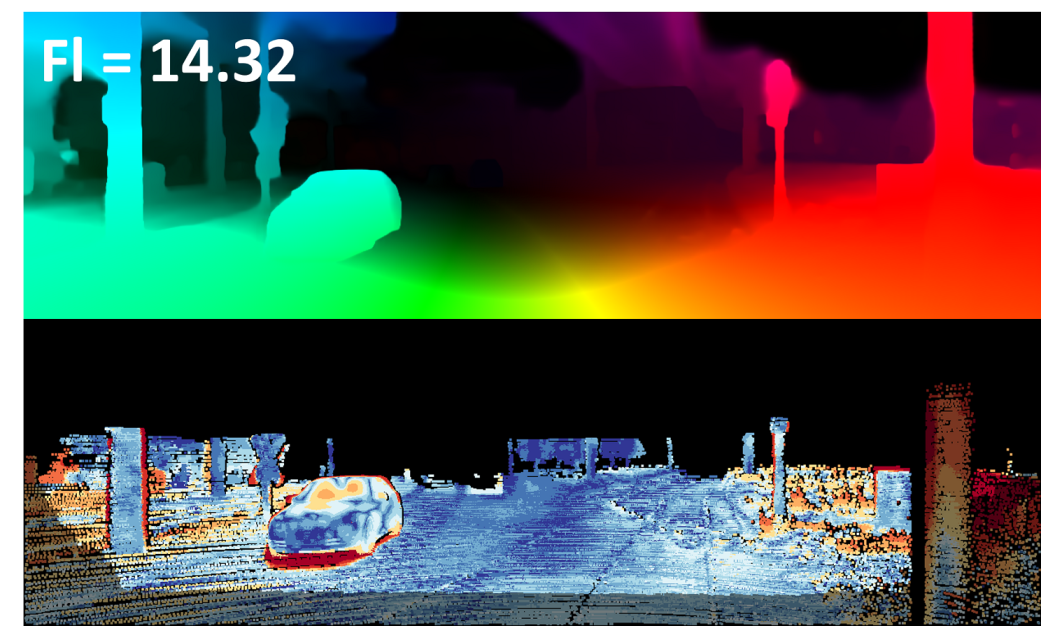}}
\hfil
\subfloat[Ours]{\includegraphics[width=1.42in]{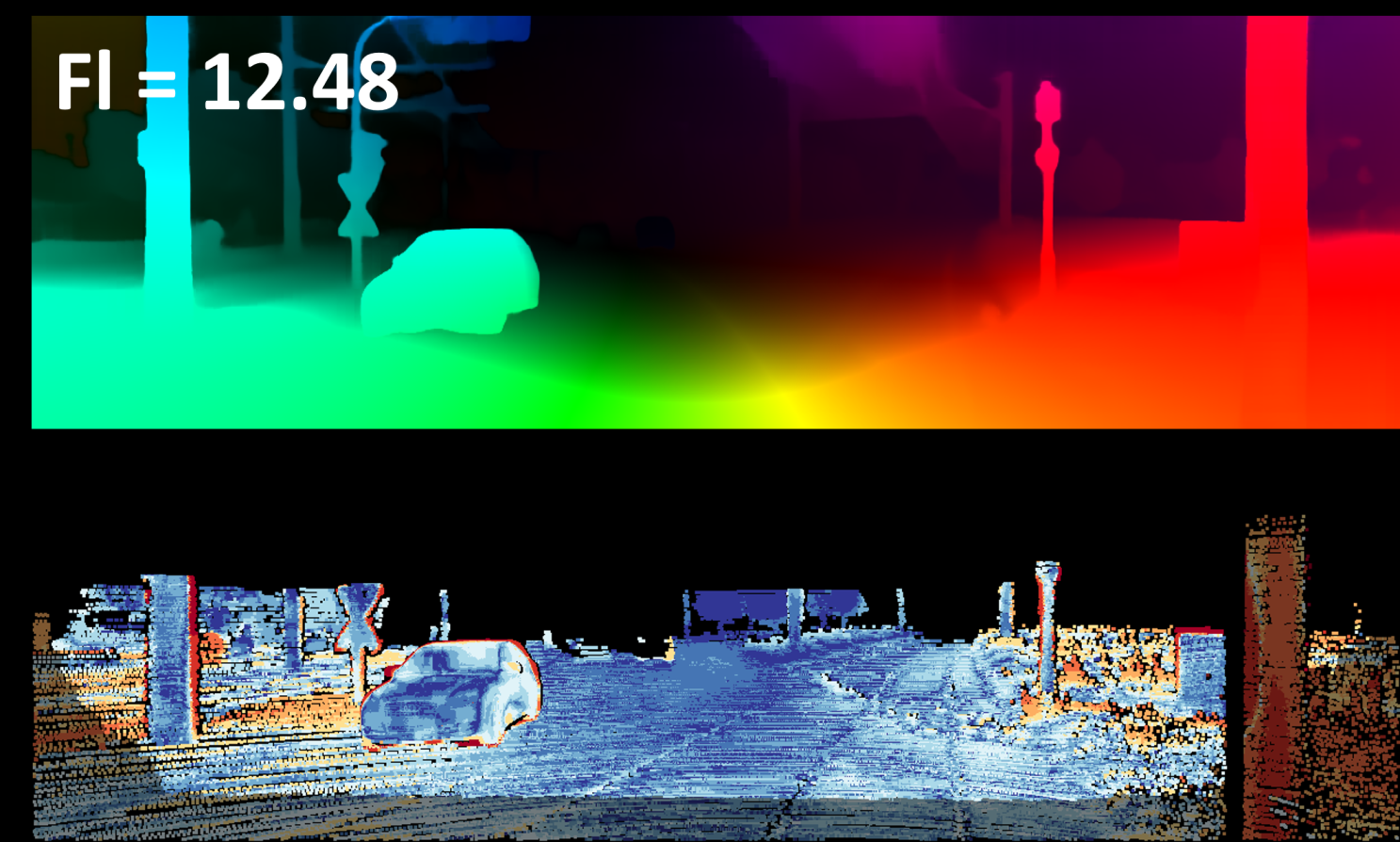}}
\caption{Unsupervised qualitative results on the KITTI-2015 test. All methods are trained without real labels on KITTI. We use the examples of the methods that are public on the KITTI Vision Benchmark. From top to bottom: predicted optical flow and error maps (sample frame \#000016).}
\label{fig:example-optical-flow2}
\end{figure*}

\begin{figure*}[t]
\centering
\subfloat[Frame 1]{\includegraphics[width=1.7in]{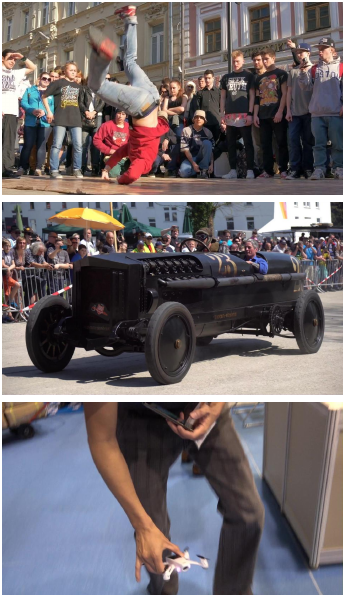}}
\hfil
\subfloat[C+T]{\includegraphics[width=1.7in]{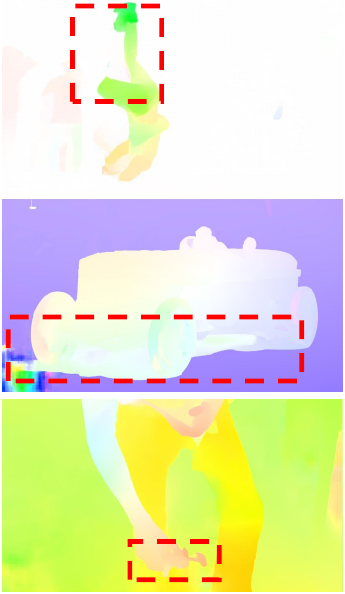}}
\hfil
\subfloat[C+T+S+K+H]{\includegraphics[width=1.7in]{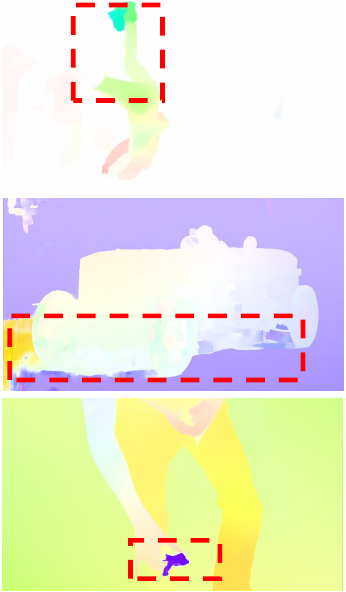}}
\hfil
\subfloat[FA-Flow Dataset]{\includegraphics[width=1.7in]{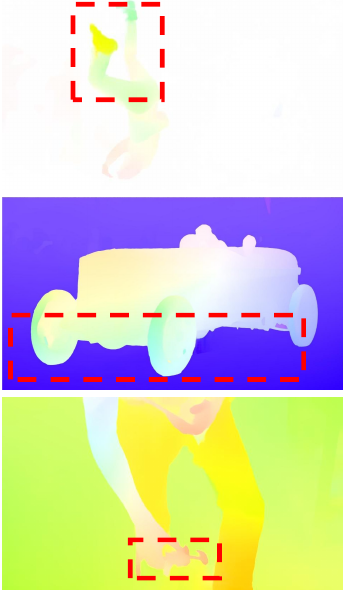}}
\caption{Qualitative results of predicted optical flow by models trained on different datasets. We show the predicted optical flow from three models here. From top to bottom: RAFT \cite{teed2020raft}, SEA-RAFT \cite{wang2024sea}, and FlowFormer++ \cite{shi2023flowformer++}. Frames are from DAVIS dataset.}
\label{fig:raft-optical-flow}
\end{figure*}

\subsection{Comparison with Synthetic Datasets}

As shown in Table \ref{tab:finetune-vk}, to validate the generalization capability of our FA-Flow Dataset, we compare it with several supervised methods trained on synthetic datasets. In this setup, models are pre-trained on ``C+T" and then fine-tuned on the dataset synthesized against KITTI 15. One such dataset is Virtual KITTI 2, which employs state-of-the-art game engines to render optical flow datasets. The setup of Virtual KITTI 2 is meticulously designed to align with KITTI 15, thereby training optical flow estimation models to improve the performance of the real images from KITTI 15. As shown in Table \ref{tab:finetune-vk}, our FA-Flow Dataset outperforms the previous best methods when compared with synthetic datasets.

We further validate our model on various real-world datasets with in-domain training data, as shown in Table \ref{tab:finetune-main}. For supervised methods, the models are pre-trained on ``C+T+S+K+H" following previous works, which indicates a mixture of FlyingChairs, FlyingThings, Sintel, FlyingThings3D clean pass, KITTI and HD1K. This training combination indicates the state-of-the-art training process. Then, models are fine-tuned on the training set of KITTI 15 and tested on the test set by uploading the estimated optical flow to the official benchmark. In this table, we train models on our mixed datasets with the generated optical flow as labels and then fine-tuned models on the labeled KITTI 15 training set. As shown in Table \ref{tab:finetune-main}, our FA-Flow Dataset demonstrates a remarkable improvement.

\subsection{Comparison with Unsupervised Methods}

Table \ref{tab:finetune-main} presents the results of a comparative analysis of various unsupervised learning methods on the KITTI 15 test set. In this experiment, all methods were trained without access to optical flow labels from the KITTI 15 training set, relying solely on the images. This setup ensures a fair comparison across different methods, as none of the models had access to ground truth flow data. Our model follows the same training protocol and is trained on our FA-Flow Dataset. As shown in the table, it outperforms other methods by a notable margin. Specifically, our model achieves a Fl-all score of $4.91$, which is much lower than the closest competitor. For background and foreground flow estimation, Flow-Anything also achieves state-of-the-art performance with Fl-bg and Fl-fg scores, respectively. These results also demonstrate the superior generalization of our model when trained using only images from the target domain, highlighting its effectiveness in unsupervised optical flow estimation.

\noindent\textbf{Qualitative Comparison.} To further verify the generalization of our proposed method in the real world, we show the results of unsupervised qualitative comparisons. In this setting, real optical flow labels from KITTI 15 are not allowed to be used. The evaluation is performed on the non-public KITTI 15 official test benchmark. We provide a qualitative comparison with previous models with the best performance on KITTI 15 in Figures \ref{fig:example-optical-flow} and \ref{fig:example-optical-flow2}. The End-Point Error (EPE) is depicted in the top-left corner. Our model consistently produces superior optical flow estimation results compared to previous models, demonstrating the effectiveness of our dataset generation method. We compare the optical flow predictions and error maps on two representative sample frames. Note that only the first $20$ samples on the test benchmark are available. The models evaluated include several unsupervised methods, which are publicly available on the official benchmark. Neither our model nor the comparison models is trained using the labels provided by KITTI. Our model achieves a lower EPE and more accurate optical flow estimates, as evidenced by the error maps. Besides, our model preserves clearer details, such as shading, the edges of objects, and fine structures. 

\section{Discussion}

To verify the effectiveness of the proposed Flow-Anything, we discuss the performance of models trained on generated datasets with different settings. All the experiments are conducted on KITTI datasets as only KITTI datasets contain real-world images and optical flow. These experiments are more reflective of how the model performs in the real world. Evaluations are performed on the training set as in previous works. Because the test set on the official benchmark is not publicly available and has very few chances to commit. In addition, we further validate the generalization of models trained with our synthetic data by leveraging frame-to-frame optical flow predictions and evaluating them on point tracking datasets and video editing benchmarks.

\begin{figure}[t]
\begin{center}
    \includegraphics[width=1.0\linewidth]{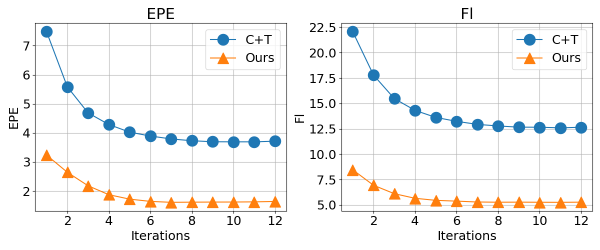}
    \end{center}
    \caption{Optical flow estimation performance on KITTI 15 with different number of refinements at inference time.}
    \label{fig:inteations}
\end{figure}

\begin{table}[t]
\centering
\caption{Comparison with synthetic datasets when optical flow estimation models are trained from scratch. In this case, we only use COCO dataset to generated the training data. The results indicates only using generated data from a sliver of data real-world images for training still leads to better performance than synthetic datasets.}
\begin{tabular}{@{}cccccc@{}}
\midrule
& ~ & \multicolumn{2}{c}{KITTI 12} & \multicolumn{2}{c}{KITTI 15} \\ \cmidrule(lr){3-4} \cmidrule(lr){5-6} 
\multirow{-2}{*}{Dataset}& \multirow{-2}{*}{Quantity} & EPE $\downarrow $ & Fl $\downarrow $ & EPE $\downarrow $ & Fl $\downarrow $ \\ \midrule
 C+T \cite{ilg2017flownet} & 47K & 2.08 & 8.86 & 5.00 & 17.44  \\
\rowcolor{graycolor}  FA-Flow (COCO) & 20K & 1.59 & 6.22 & 3.68 & 11.95  \\
\rowcolor{graycolor}  FA-Flow (COCO) & 120K & \textbf{1.51} & \textbf{5.94} & \textbf{3.41} & \textbf{11.16}  \\ \midrule
\end{tabular}
\label{tab:without-pre-train}
\end{table}

\begin{table*}[t]
\centering
\caption{Evaluation results of frame-to-frame optical flow models on point tracking datasets. We report results on TAP-Vid benchmark with three datasets. AJ: Average Jaccard; $<\delta t$: Temporal alignment; OA: Overall Accuracy.}
\begin{tabular}{@{}llccccccccc@{}}
\toprule
\multirow{2}{*}{Model} & \multirow{2}{*}{Training Dataset} 
& \multicolumn{3}{c}{TAP-Vid-Kinetics} 
& \multicolumn{3}{c}{TAP-Vid-DAVIS} 
& \multicolumn{3}{c}{TAP-Vid-RGB-Stacking} \\ 
\cmidrule(lr){3-5} \cmidrule(lr){6-8} \cmidrule(lr){9-11}
& & AJ $\uparrow$ & $<\delta t$ $\uparrow$ & OA $\uparrow$
  & AJ $\uparrow$ & $<\delta t$ $\uparrow$ & OA $\uparrow$
  & AJ $\uparrow$ & $<\delta t$ $\uparrow$ & OA $\uparrow$ \\
\midrule
\multirow{3}{*}{RAFT} 
& C+T & 37.02 & 52.11 & 80.40 & 35.05 & 49.72 & 79.71 & 43.07 & 53.71 & 90.69 \\
& C+T+S+K+H & \underline{39.81} & \underline{55.02} & \underline{80.69} & \underline{36.58} & \underline{51.36} & \underline{79.89} & \underline{49.77} & \underline{60.53} & \underline{92.19} \\
 \rowcolor{graycolor} \cellcolor{white} & \textbf{FA-Flow Dataset (ours)} & \textbf{39.91} & \textbf{55.24} & \textbf{80.73} & \textbf{37.62} & \textbf{51.80} & \textbf{79.92} & \textbf{50.06} & \textbf{61.92} & \textbf{92.31} \\
\midrule
\multirow{3}{*}{SEA-RAFT} 
& Tartan+C+T & 40.41 & 54.43 & \underline{81.77} & 33.84 & 48.52 & 79.82 & \underline{52.57} & \underline{63.33} & \textbf{91.93} \\
& Tartan+C+T+S+K+H & \textbf{44.25} & \textbf{58.30} & \textbf{82.45} & \underline{41.69} & \underline{56.90} & \textbf{80.10} & 52.55 & 63.34 & \underline{91.41} \\
\rowcolor{graycolor}  \cellcolor{white} & \textbf{FA-Flow Dataset (ours)} & \underline{41.67} & \underline{56.97} & 81.17 & \textbf{42.70} & \textbf{57.71} & \underline{80.26} & \textbf{54.36} & \textbf{65.07} & \textbf{91.94} \\
\bottomrule
\end{tabular}
\label{tab:tracking}
\end{table*}

\begin{figure*}[t]
\centering
\subfloat[Tartan+C+T]{\includegraphics[width=6.9in]{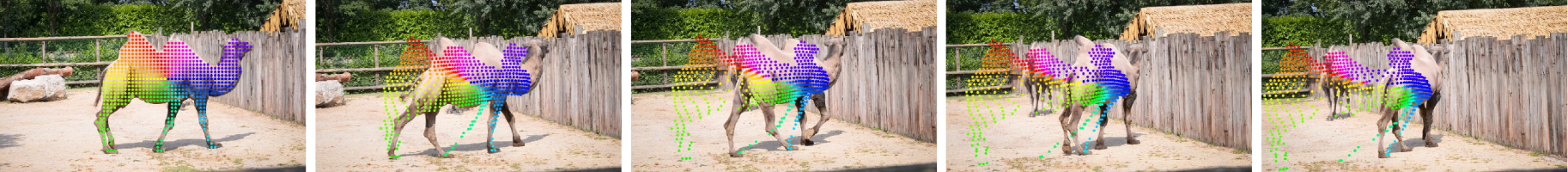}}
\hfil
\subfloat[Tartan+C+T+S+K+H]{\includegraphics[width=6.9in]{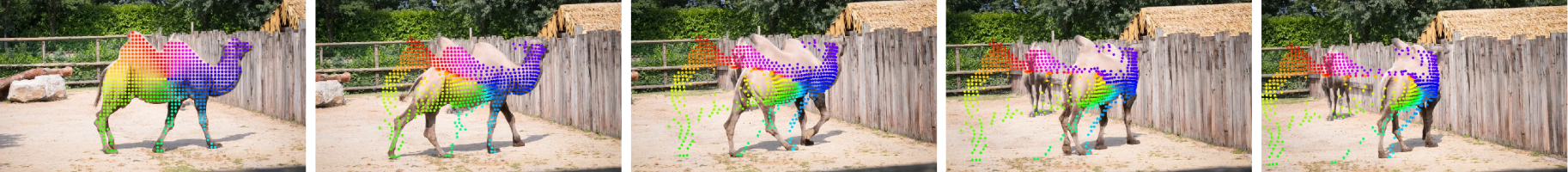}}
\hfil
\subfloat[FA-Flow Dataset]{\includegraphics[width=6.9in]{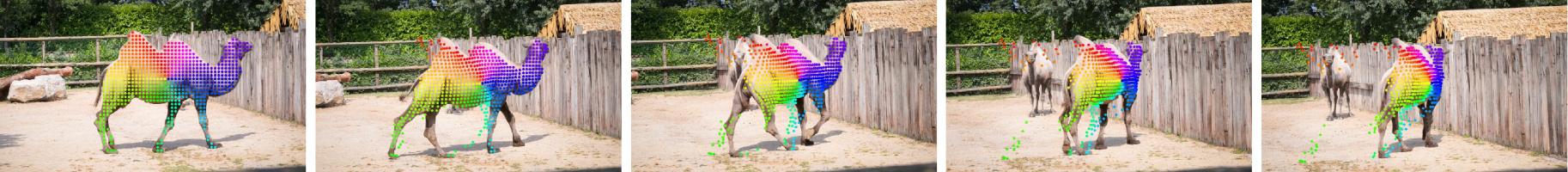}}
\caption{Qualitative results of SEA-RAFT trained on different datasets on point tracking datasets with frame-to-frame optical flow predictions.}
\label{fig:tracking-optical-flow}
\end{figure*}


\subsection{Benefits of Real-World Images for Training}

In this section, we discuss the benefits of real-world images for training optical flow estimation models, especially in comparison to synthetic data, as shown in Table \ref{tab:without-pre-train}. We compare the performance of models trained on our generated data with synthetic data under identical training settings to evaluate their effectiveness when trained from scratch. The results clearly demonstrate that models trained on datasets derived from real-world images outperform the model trained on the synthetic ``C+T" dataset when evaluated on the KITTI benchmarks. This indicates a significant domain gap between synthetic datasets and real-world applications. Even though the optical flow labels or novel frames in the generated datasets may be relatively inaccurate compared to the highly accurate labels in ``C+T", models trained on these real-world images still achieve superior performance.

We further validate the benefits of real-world images for training by evaluating optical flow estimation models trained on different datasets. To be specific, we visualize the predicted optical flow of RAFT models trained on ``C+T", ``C+T+S+K+H", and generated data using our proposed Flow-Anything, in which ``C+T+S+K+H" indicates the state-of-the-art combination of training data to date. Our proposed method leverages a large-scale dataset of real-world images for training. By contrast, ``"C+T" and ``C+T+S+K+H" predominantly rely on synthetic datasets or limited real-world data with sparse flow annotations. 

Figure \ref{fig:raft-optical-flow} illustrates the qualitative results of predicted optical flow from models trained on different datasets. The first row demonstrates the handling of missing parts in blurred images, such as the upper body of the person in the first frame. Traditional models fail to accurately capture the motion, resulting in incomplete or distorted flow fields. Model trained on our FA-Flow Dataset, however, demonstrates a superior ability to recover the motion details in these blurred regions. The second row focuses on another challenge of estimating optical flow in the presence of shadows. Our model mitigates these issues by correctly distinguishing between the actual object motion and shadow-induced artifacts. In the third row, the models' ability to capture the motion of small objects and accurately predict motion direction is evaluated. The edges of small objects, such as the fingers in the last frame, are often lost or poorly estimated by other models. Furthermore, the predicted motion direction by ``C+T+S+K+H" can be incorrect, as evidenced by the misaligned flow vectors. Our model excels in preserving the details of small objects and provides a precise motion direction, showcasing its robustness and fine-grained accuracy.

Figure \ref{fig:inteations} also shows the quantitative results of SEA-RAFT trained on different datasets with different numbers of refinement steps. The models trained on both ``C+T" and our generated data start with a high EPE and Fl at the first iteration and show a steady decrease as the number of iterations increases. The results indicate that our method consistently outperforms the ``C+T" model across all iterations. The lower EPE values demonstrate that SEA-RAFT trained our FA-Flow Dataset achieves better accuracy in predicting optical flow. Another observation from the results is that our proposed FA-Flow Dataset leads to the early stabilization of both EPE and Fl metrics. This early stabilization implies that our model reaches its optimal performance much faster than the ``C+T" model. The early stabilization of EPE and Fl in our proposed FA-Flow Dataset also translates to reduced inference computation.

\subsection{Performance on Point Tracking Datasets}

To further validate the generalization ability of models trained on our FA-Flow Dataset, we evaluate their performance on point tracking benchmarks using frame-to-frame optical flow predictions. As shown in Table~\ref{tab:tracking}, we report results on the TAP-Vid benchmark \cite{doersch2022tap} across three datasets: TAP-Vid-Kinetics, TAP-Vid-DAVIS, and TAP-Vid-RGB-Stacking. We follow the standard evaluation protocol and use RAFT and SEA-RAFT as base models for comparison. Our models are trained solely on the FA-Flow Dataset without any fine-tuning on the TAP-Vid datasets.

The results show that models trained on FA-Flow achieve competitive or superior performance compared to those trained on traditional combinations of real and synthetic datasets (e.g., C+T+S+K+H). Notably, SEA-RAFT trained on FA-Flow achieves the best performance on TAP-Vid-DAVIS and TAP-Vid-RGB-Stacking, demonstrating strong robustness and generalization in diverse and dynamic scenes.

Furthermore, as illustrated in Figure \ref{fig:tracking-optical-flow}, we provide qualitative visualizations of long-term tracking results using only frame-to-frame predictions from SEA-RAFT. Despite lacking explicit temporal modeling, the model trained on our synthetic data maintains consistent tracking over time. These results highlight the diversity and effectiveness of our generated training data in supporting real-world, long-term motion understanding.

\begin{table}[t]
\centering
\caption{Ablation experiments on our main modules. Settings used are marked with a checkmark. ``DO" indicates dynamic objects. ``OAI" indicates object aware inpainting. ``MO" indicates multiple objects.}
\begin{tabular}{@{}ccccccc@{}}
\toprule
 \multirow{2}{*}{DO} & \multirow{2}{*}{OAI} & \multirow{2}{*}{MO} & \multicolumn{2}{c}{KITTI 12} & \multicolumn{2}{c}{KITTI 15} \\ \cmidrule(lr){4-5} \cmidrule(lr){6-7} 
 &  &  & EPE $\downarrow $ & Fl $\downarrow $ & EPE $\downarrow $ & Fl $\downarrow $ \\ \midrule
  \ding{53} & \ding{53} & \ding{53} & 1.19 & 4.34 & \{2.47\} & \{8.16\} \\
  \ding{51} & \ding{53} & \ding{53} & 1.29 & 4.48 & \{1.80\} & \{7.01\} \\ 
  \ding{51} & \ding{51} & \ding{53} & 1.24 & 4.31 & \{1.85\} & \{7.01\}  \\
  \ding{51} & \ding{51} & \ding{51} & \textbf{1.17} & \textbf{4.29} & \{\textbf{1.79}\} & \{\textbf{6.98}\}  \\ \bottomrule
\end{tabular}
\label{tab:ablations}
\end{table}

\subsection{Ablations on Main Modules}
\label{sub_ablation}

To analyze the impact of different components in our proposed Flow-Anything for optical flow data generation, we conduct a series of ablation studies on object motion, Depth-Aware Inpainting, and multiple objects, as shown in Table \ref{tab:ablations}.  ``Multiple objects" indicates the presence of multiple moving objects in each image. In these experiments, we generate novel view images from the KITTI 15 training set to train the RAFT model and evaluate the performance on both the KITTI 12 and KITTI 15 training sets. By incrementally adding components of our method, we can measure the impact of each factor, as shown in Table \ref{tab:ablations}. The first row shows the performance achieved by generating novel view images without dynamic objects, assuming that optical flow results solely from camera motion. The second row introduces dynamic objects into the scene. We observe a significant improvement in the EPE metrics, particularly in the KITTI 15 dataset, where the presence of dynamic objects increases the complexity of the optical flow estimation task. The addition of Depth-Aware Inpainting, as shown in the third row, further enhances the model's performance by providing a more accurate representation of occluded regions, which is critical for generating realistic optical flow data. Finally, the fourth row incorporates multiple objects, demonstrating the robustness of our method in handling complex scenes with multiple moving objects. This comprehensive setup yields the best results across all evaluated metrics, particularly in the KITTI 15 dataset, where both the EPE and F1 scores are significantly improved.

Figure \ref{fig:incremental} provides a detailed illustration of the incremental effects of Flow-Anything with and without Object-Independent Volume Rendering and Depth-Aware Inpainting. As shown in Figures \ref{11} and \ref{22}, camera motion alone does not capture the complex optical flow shown in real-world scenes. To address these limitations, we introduce the Object-Independent Volume Rendering module, which ensures that dynamic objects have separate motions from the background. This module helps to capture the complex movements within the scene more accurately, as shown in Figures \ref{33}. Additionally, we apply the Depth-Aware Inpainting module to enhance the realism further by filling in masked areas caused by object motions, effectively handling occlusions, and improving the continuity of the flow fields, as shown in Figures \ref{44}.

\section{Applications}

In this section, we demonstrate the effectiveness of models trained on our FA-Flow Dataset in several applications. We mainly show the effects of optical flow models trained on two types of datasets. First, the optical flow model used by default in downstream methods, in which RAFT trained on ``C+T+S+K+H" is used. Then, the RAFT trained on generated data from real-world images using our Flow-Anything, which we label as FA-Flow Dataset. Experiments are conducted on representative tasks, including video inpainting, unsupervised video segmentation, space-time view synthesis, and video editing. In these tasks, optical flow plays a crucial role in each of these tasks.

\begin{figure}[t]
\centering
\subfloat[Frames]{\includegraphics[width=1.15in]{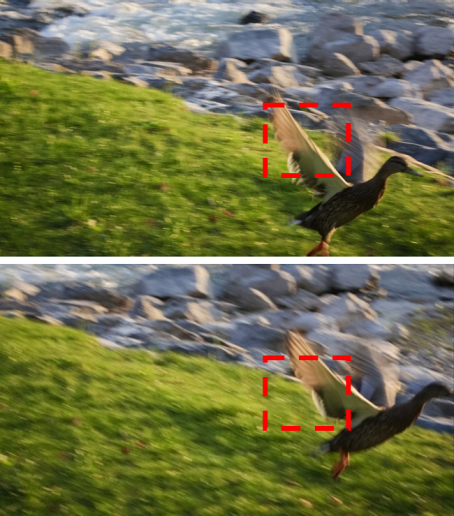}}
\hfill
\subfloat[C+T+S+K+H]{\includegraphics[width=1.15in]{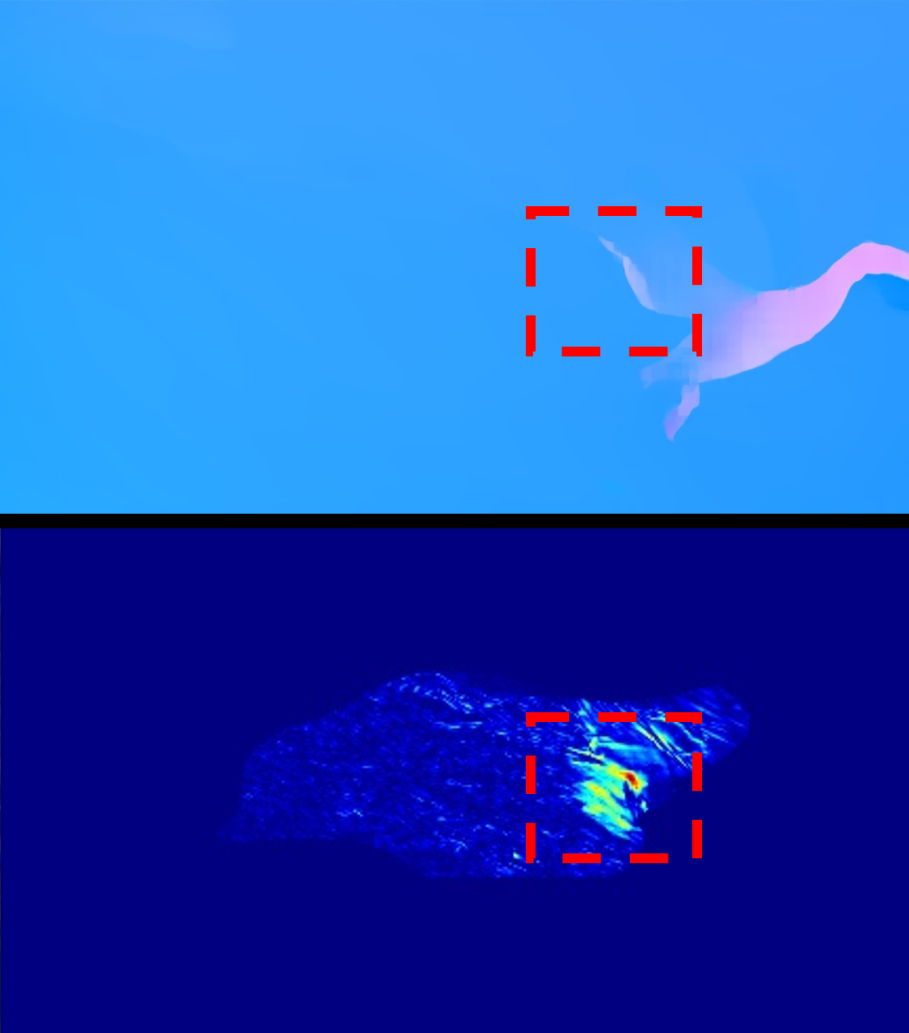}}
\hfill
\subfloat[FA-Flow Dataset]{\includegraphics[width=1.15in]{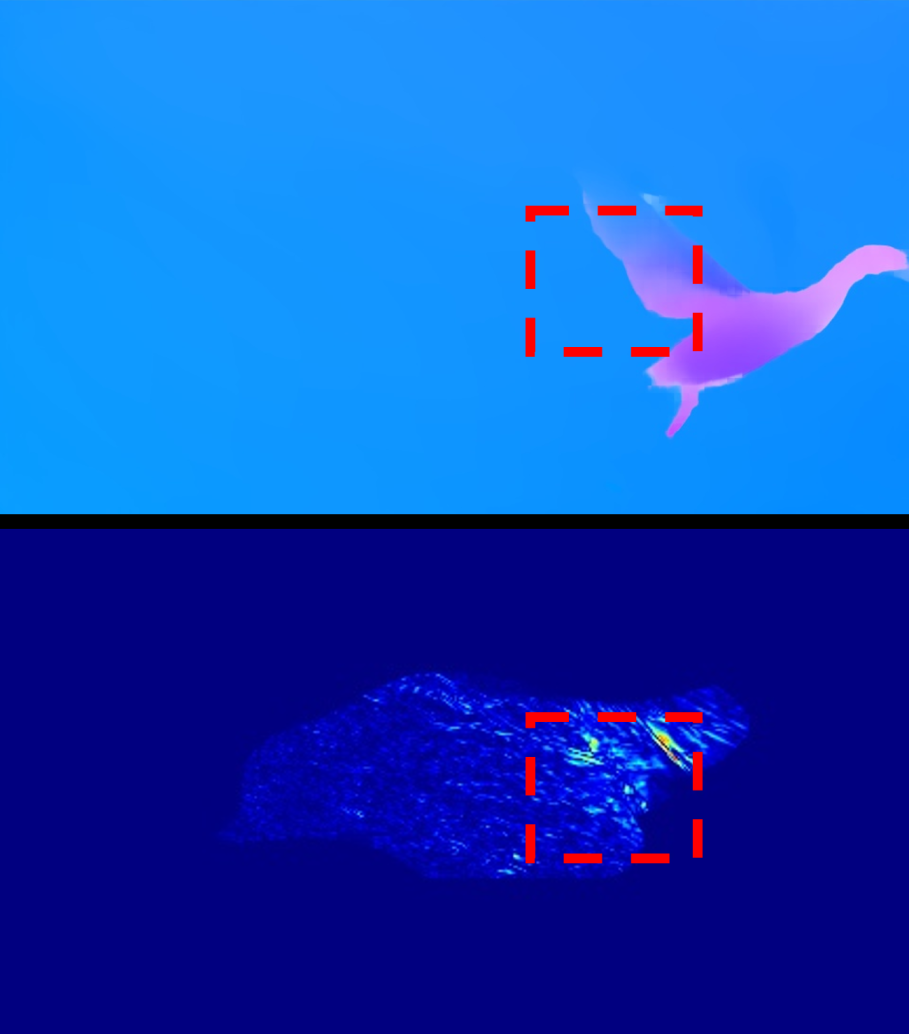}}
\caption{Qualitative comparison of ProPainter using RAFT trained on ``C+T+S+K+H" or our FA-Flow Dataset for video inpainting.}
\label{fig:inpainting-example}
\end{figure}

\begin{table}[t] 
\centering
\caption{Quantitative results for the inpainting task are presented. We use ProPainter for inpainting and evaluate the models on the YouTube-VOS and DAVIS datasets. Optical flows, used as inputs, are predicted from RAFT models trained on the "C+T+S+K+H" combination or our FA-Flow Dataset, respectively. The datasets used to train the optical flow models are indicated in the "Flow Dataset" column.}
\begin{tabular}{@{}lccc@{}}
\toprule
Validation Dataset & Flow Dataset & \multicolumn{1}{c}{PSNR $\uparrow$} & \multicolumn{1}{c}{SSIM $\uparrow$} \\ \midrule
 & C+T+S+K+H & 33.96 & 0.9718  \\
 \rowcolor{graycolor} \cellcolor{white} \multirow{-2}{*}{YouTube-VOS \cite{xu2018youtube}}  &  FA-Flow Dataset & \textbf{34.04} & \textbf{0.9726} \\ \midrule
  & C+T+S+K+H & 34.17 & 0.9771 \\
   \rowcolor{graycolor} \cellcolor{white} \multirow{-2}{*}{DAVIS \cite{perazzi2016benchmark}} & FA-Flow Dataset & \textbf{34.30} & \textbf{0.9774} \\ \bottomrule
\end{tabular}%
\label{tab:video-enhancement-comparison}
\end{table}

\subsection{Video Inpainting}

Video inpainting involves removal of objects from video sequences while maintaining temporal coherence. This technique is crucial in various applications, where unwanted objects need to be removed without leaving noticeable artifacts or inconsistencies. Flow-based propagation is one mainstream mechanism in video inpainting to ensure that the background and remaining content in the video appear natural and coherent across frames. 

To validate the performance of RAFT trained on our FA-Flow Dataset in video inpainting, we conducted experiments on a variety of video sequences with different objects removed. We compared RAFT trained on our FA-Flow Dataset with the one trained on ``C+T+S+K+H" using ProPainter \cite{zhou2023propainter}, which indicates the state-of-the-art video inpainting model using optical flow. Figure \ref{fig:inpainting-example} illustrates a series of frames from a video sequence where objects are removed using our model. The results highlight the ability of our model to maintain temporal coherence and produce visually appealing backgrounds. Additionally, we used quantitative metrics such as Peak Signal-to-Noise Ratio (PSNR) and Structural Similarity Index (SSIM) to evaluate the performance. As shown in Table \ref{tab:video-enhancement-comparison}, ProPainter with our model outperforms the one using RAFT on both YouTube-VOS and DAVIS datasets, achieving higher PSNR and SSIM scores, and lower VFlD.

\begin{table}[t]
\centering
\caption{Quantitative results on unsupervised video segmentation with TokenCUt and TMO. mIOU is reported. We set the max frames for graph construction of TokenCut to 50 for memory efficiency.}
\begin{tabular}{@{}lccc@{}}
\toprule
\multirow{2}{*}{Method} & \multirow{2}{*}{Flow Dataset} & \multicolumn{2}{c}{Validation Dataset} \\ \cmidrule(lr){3-4} 
& & DAVIS & FBMS \\ \midrule
& C+T+S+K+H & 76.17 & 66.10 \\
\rowcolor{graycolor} \cellcolor{white} \multirow{-2}{*}{TokenCut \cite{wang2023tokencut}} & FA-Flow Dataset & \textbf{76.82} & \textbf{67.45} \\ \midrule
& C+T+S+K+H & 85.60 & 79.90 \\
\rowcolor{graycolor} \cellcolor{white} \multirow{-2}{*}{TMO \cite{cho2023treating}} & FA-Flow Dataset & \textbf{86.16} & \textbf{80.04} \\ \bottomrule
\end{tabular}
\label{tab:uvos}
\end{table}

\begin{figure}[t]
\centering
\begin{minipage}[t]{0.49\textwidth}
    \centering
    \subfloat[C+T+S+K+H]{\includegraphics[width=3.5in]{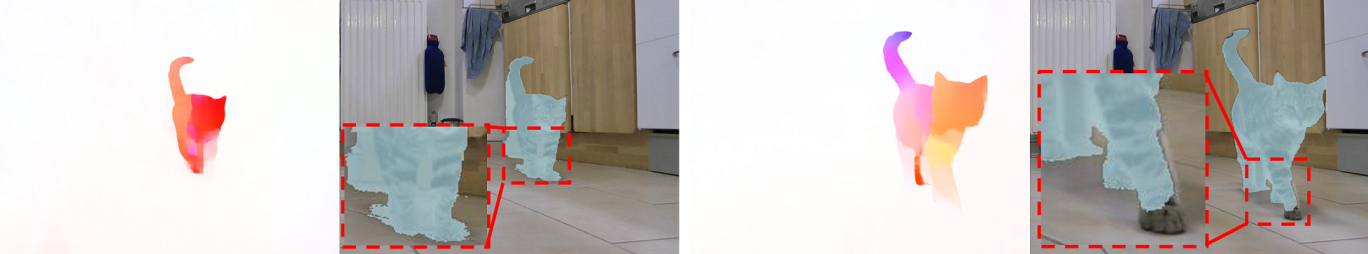}}
    \hfil
    \subfloat[FA-Flow Dataset]{\includegraphics[width=3.5in]{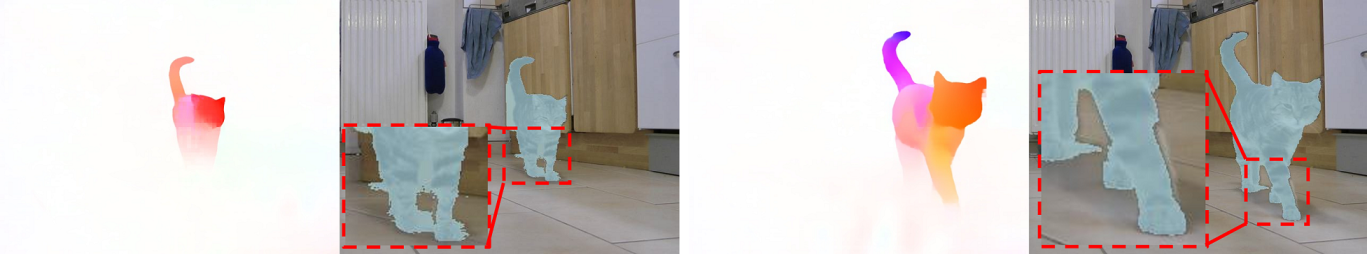}}
    \hfill
    \caption{Qualitative comparison of TokenCut using RAFT trained on ``C+T+S+K+H" or our FA-Flow Dataset for unsupervised video segmentation.}
    \label{fig:samples-segmentation}
\end{minipage}

\vspace{0.4cm}

\begin{minipage}[t]{0.49\textwidth}
    \centering
    \subfloat[C+T+S+K+H]{\includegraphics[width=3.5in]{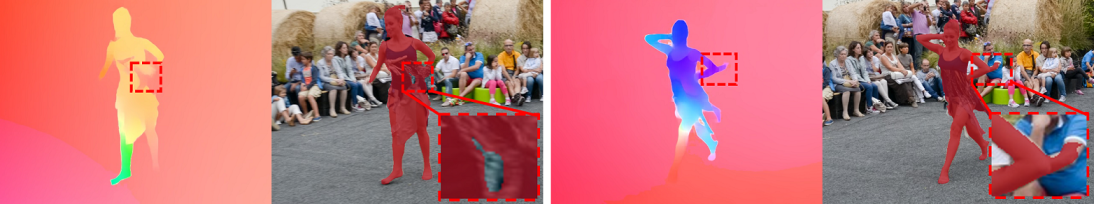}}
    \hfil
    \subfloat[FA-Flow Dataset]{\includegraphics[width=3.5in]{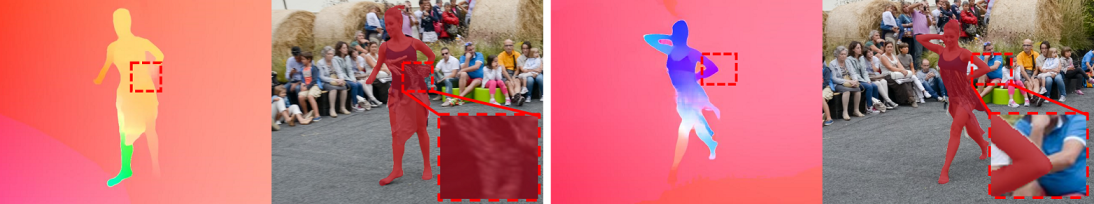}}
    \caption{Qualitative comparison of TMO using RAFT trained on ``C+T+S+K+H" or our FA-Flow Dataset for unsupervised video segmentation.}
    \label{fig:samples-segmentation2}
\end{minipage}
\end{figure}

\begin{table*}[ht]
\centering
\caption{Quantitative results on space-time view synthesis. We report the PSNR $\uparrow$ / LPIPS $\downarrow$ results with comparisons on Dynamic Scene dataset.}
\begin{tabular}{@{}lcccccccc@{}}
\toprule
Method & Flow Dataset & Jumping & Skating & Truck & Umbrella & Balloon1 & Balloon2 & Playground \\ \midrule
\multirow{2}{*}{DyNeRF \cite{gao2021dynamic}} & C+T+S+K+H & 24.57 / \textbf{0.091} & 32.32 / 0.037 & 26.22 / 0.077 & 23.39 / 0.141 & \textbf{22.38} / 0.105 & 26.81 / 0.053 & 24.34 / 0.084 \\
 \rowcolor{graycolor} \cellcolor{white} & FA-Flow Dataset  & \textbf{24.74} / 0.094 & \textbf{32.64} / \textbf{0.036} & \textbf{26.26} / \textbf{0.076} & \textbf{23.52} / \textbf{0.138} & 22.36 / \textbf{0.104} & \textbf{26.88} / \textbf{0.052} & \textbf{24.40} / \textbf{0.083} \\ 
\midrule
\multirow{2}{*}{NSFF \cite{li2021neural}} & C+T+S+K+H & 26.99 / 0.057 & 33.01 / 0.021 & 32.08 / 0.028 & 24.45 / \textbf{0.096} & \textbf{24.15} / 0.066 & 29.51 / 0.035 & 24.61 / \textbf{0.064} \\
 \rowcolor{graycolor} \cellcolor{white} & FA-Flow Dataset & \textbf{27.05} / \textbf{0.055} & \textbf{34.50} / \textbf{0.020} & \textbf{32.17} / \textbf{0.026} & \textbf{24.50} / 0.100 & \textbf{24.15} / \textbf{0.065} & \textbf{30.98} / \textbf{0.029} & \textbf{24.64} / \textbf{0.064} \\
 \bottomrule
\end{tabular}
\label{tab:dync-nerf}
\end{table*}

\subsection{Unsupervised Video Segmentation}

Unsupervised video object segmentation aims to segment a target object in the video without a ground truth mask in the initial frame. This challenging task requires extracting features for the most salient objects within a video sequence using motion information such as optical flow. Methods should provide a set of object candidates with no overlapping pixels that span through the whole video sequence. This set of objects should contain at least the objects that capture human attention when watching the whole video sequence. Optical flow provides critical motion cues that help in identifying and tracking objects.

In our experiments, we focus on two representative methods for unsupervised video segmentation: TokenCut \cite{wang2023tokencut} and TMO \cite{cho2023treating}. We compare the downstream performance of RAFT trained on our FA-Flow Dataset with the default ``C+T+S+K+H" used in these two methods across two evaluation datasets: DAVIS and FBMS. The mean Intersection over Union (mIOU) metric is used to evaluate the performance. The quantitative results are shown in Table \ref{tab:uvos}. The results show that our model outperforms RAFT train on synthetic datasets in all cases, indicating the superiority of our model in capturing motion information for video segmentation. The results show the efficacy of Flow-Anything in improving the segmentation accuracy of unsupervised methods. We also provide a qualitative comparison of the segmentation results using TokenCut and TMO with both RAFT and Flow-Anything as optical flow estimators, as shown in Figures \ref{fig:samples-segmentation} and \ref{fig:samples-segmentation2}. As observed, our model consistently delivers more precise and coherent segmentation masks.

\subsection{Space-Time View Synthesis}

\begin{figure}[t]
\centering
\begin{minipage}[t]{0.49\textwidth}
    \centering
    \subfloat[Frames]{\includegraphics[width=1.15in]{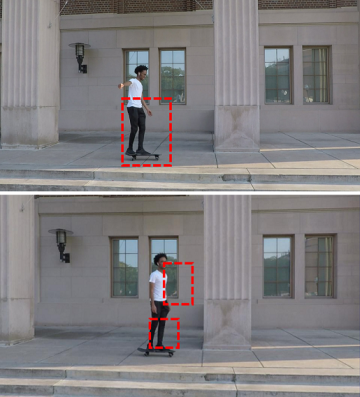}}
    \hfill
    \subfloat[C+T+S+K+H]{\includegraphics[width=1.15in]{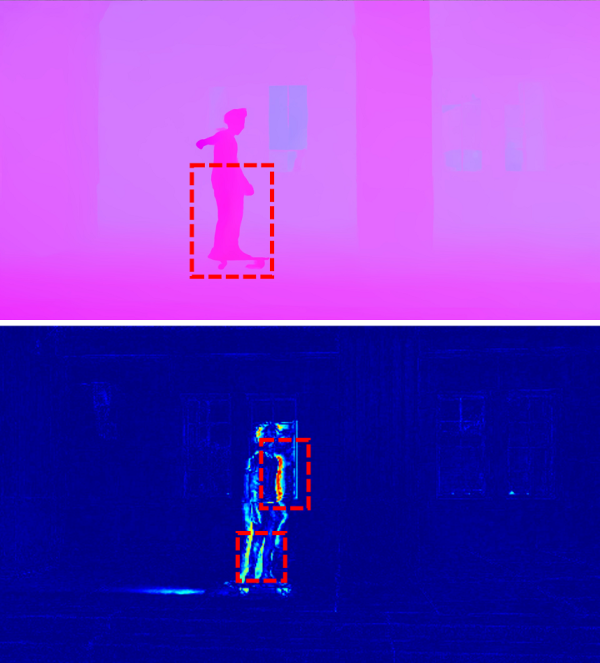}}
    \hfill
    \subfloat[FA-Flow Dataset]{\includegraphics[width=1.15in]{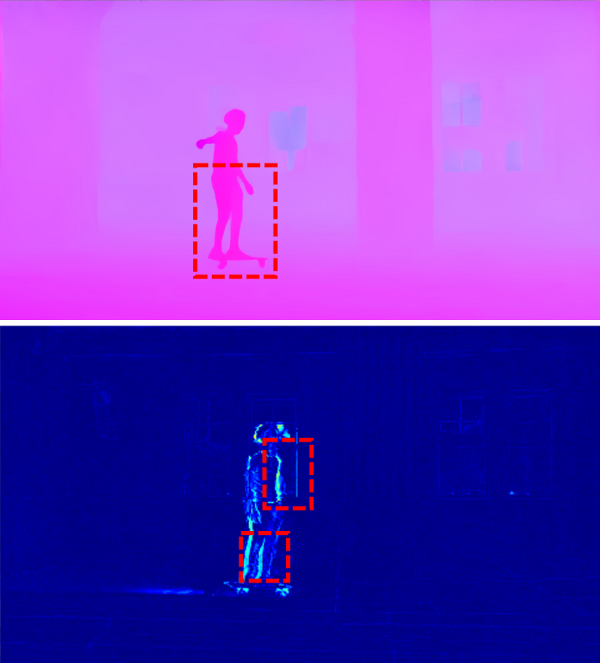}}
    \caption{Qualitative comparison of DyNeRF using RAFT trained on ``C+T+S+K+H" and our FA-Flow Dataset for space-time view synthesis.}
    \label{fig:dynerf-example}
\end{minipage}
\hfill
\begin{minipage}[t]{0.49\textwidth}
    \centering
    \subfloat[Frames]{\includegraphics[width=1.15in]{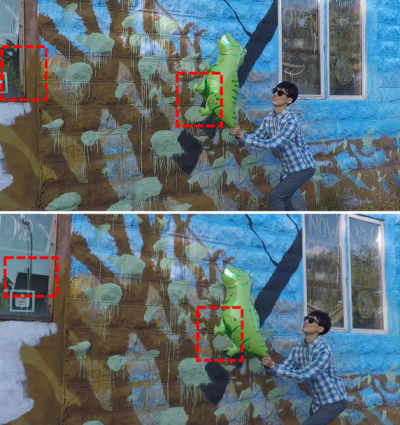}}
    \hfill
    \subfloat[C+T+S+K+H]{\includegraphics[width=1.15in]{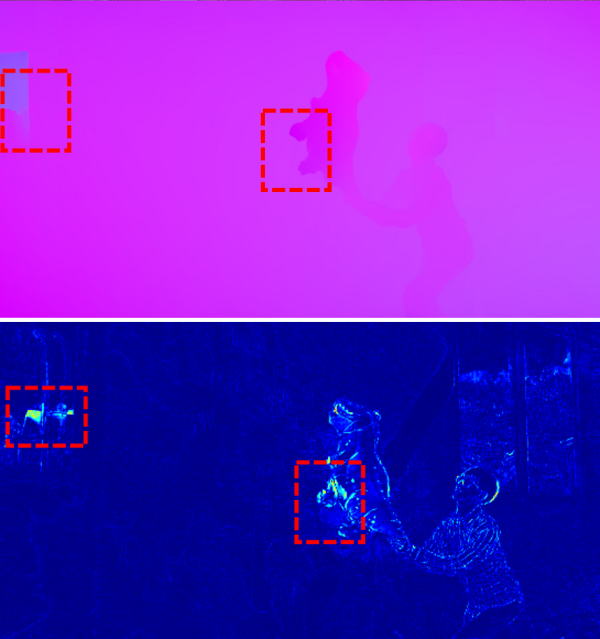}}
    \hfill
    \subfloat[FA-Flow Dataset]{\includegraphics[width=1.15in]{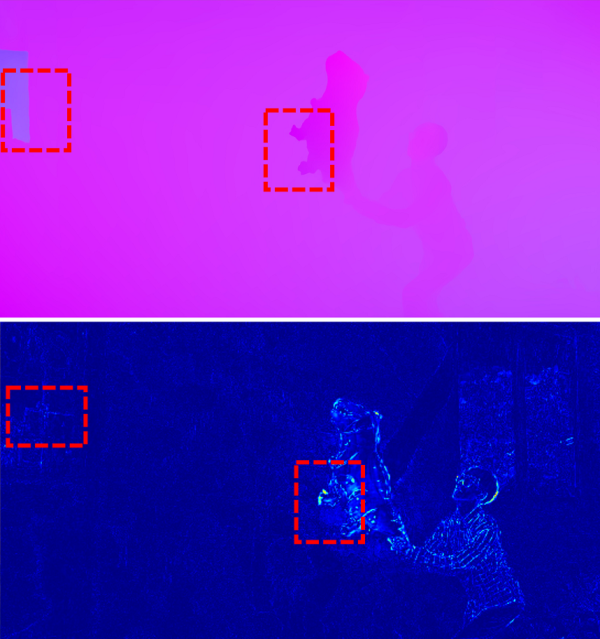}}
    \caption{Qualitative comparison of NSFF using RAFT trained on ``C+T+S+K+H" and our FA-Flow Dataset for space-time view synthesis.}
    \label{fig:nsff-example}
\end{minipage}
\end{figure}

Space-Time View Synthesis involves generating novel views of a scene over time. This method captures the temporal evolution of a scene, allowing for the reconstruction of dynamic environments from different perspectives. Optical flow estimation plays a pivotal role in enhancing the accuracy and realism of space-time view synthesis. Optical flow estimation helps in temporal consistency and accurately tracking the movement of objects. To evaluate the effectiveness of our model, we conducted experiments comparing two representative methods, DyNeRF \cite{gao2021dynamic} and NSFF \cite{li2021neural}, using RAFT and our model for optical flow estimation. For fair comparison, checkpoints saved at $300,000$ and $250,000$ iterations are used for evaluation. Table \ref{tab:dync-nerf} shows the results of these experiments, reporting the average peak signal-to-noise ratio (PSNR) and learned perceptual image patch similarity (LPIPS) scores in the Dynamic Scene dataset. Higher PSNR values and lower LPIPS values indicate better performance in terms of image quality and perceptual similarity, respectively. As shown in Table\ref{tab:dync-nerf}, using our model outperforms RAFT trained on synthesis datasets across most scenes in terms of both PSNR and LPIPS. 

We also visualize the rendered results using default optical flow models and our model in Figures \ref{fig:dynerf-example} and \ref{fig:nsff-example}. Our model achieves significantly better optical flow estimation on the edges of dynamic objects and the motion of reflections on glass surfaces. This improved optical flow estimation translates to better novel view synthesis results, yielding more accurate reconstructions.

\begin{figure}[t]
\centering
\subfloat[C+T+S+K+H]{\includegraphics[width=3.3in]{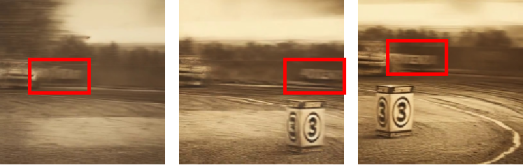}}
\hfil
\subfloat[FA-Flow Dataset]{\includegraphics[width=3.3in]{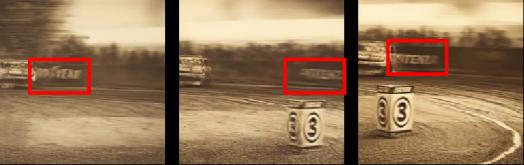}}
\caption{Qualitative results of FLATTEN with SEA-RAFT trained on different datasets for video editing.}
\label{fig:editing-optical-flow}
\end{figure}

\begin{table}[t]
\centering
\caption{Quantitative results on Video Editing with FLATTEN. We report the CLIP scores and Pick Score on the TGVE-DAVIS dataset.}
\begin{tabular}{@{}lcccc@{}}
\toprule
Method & Flow Dataset & CLIP-F & PickScore & CLIP-T \\ \midrule
\multirow{2}{*}{FLATTEN} & C+T+S+K+H & 0.915 & 20.73 & 27.78 \\ 
\rowcolor{graycolor} \cellcolor{white}  & FA-Flow Dataset & \textbf{0.916} & \textbf{20.79} & \textbf{27.95} \\ \bottomrule
\end{tabular}
\label{tab:flatten}
\end{table}

\subsection{Video Editing}

Video editing applications, such as object-driven manipulation and stylization, often rely on accurate optical flow to maintain temporal coherence and structure consistency across frames. In this section, we evaluate the effectiveness of our FA-Flow Dataset in the context of video editing using FLATTEN~\cite{cong2024flatten}, a recent representative framework that leverages optical flow for temporal consistency.

To assess the impact of our synthetic data, we replace this component with SEA-RAFT trained on ``C+T+S+K+H" and our FA-Flow Dataset, separately. For fair comparison, we follow the official FLATTEN repository’s default settings with a guidance scale of 20, fixed random seed, and no use of negative prompts. Quantitative results on the TGVE-DAVIS \cite{wu2023tune} benchmark are reported in Table~\ref{tab:flatten}, where we evaluate the edited results using CLIP-based metrics, including CLIP-F, CLIP-T, and PickScore. The differences between the two training settings are relatively minor, which is expected given the strong temporal smoothing and image refinement strategies employed in the FLATTEN framework. These mechanisms tend to minimize the effect of moderate flow quality differences.

As shown in Figure~\ref{fig:editing-optical-flow}, qualitative comparisons also reveal that in certain cases, the model trained with FA-Flow yields more consistent results. Specifically, our model better preserves textual elements and fine structures across frames, suggesting its advantage in scenarios where maintaining semantic fidelity is crucial.

\section{Conclusion}

In this paper, we introduce a novel framework, Flow-Anything, for generating large-scale optical flow datasets, addressing the challenges of image realism and motion realism. Our method introduces an MPI-based rendering pipeline that produces realistic images and corresponding optical flow from novel views. This pipeline employs volume rendering to mitigate image artifacts, resulting in higher quality. We further enhance motion realism with our Object-Independent Volume Rendering module, which separates dynamic objects from the static background, improving the accuracy of object motion. Additionally, the Depth-Aware Inpainting module addresses unnatural occlusions caused by object motion, ensuring smoother and more realistic results. By converting single-view images into 3D representations using advanced monocular depth estimation networks and simulating realistic 3D motion under a virtual camera, our method effectively bridges the gap between synthetic data and real-world applications. 

Flow-Anything demonstrates superior performance on real-world datasets, outperforming both unsupervised and supervised methods commonly used for training learning-based models. Our framework trains robust optical flow estimation models on large-scale data generated from diverse single-view images across various domains, achieving state-of-the-art performance on standard benchmarks. Furthermore, Flow-Anything enhances the performance of downstream tasks, including video inpainting, unsupervised video segmentation, and space-time view synthesis. In conclusion, Flow-Anything serves as a competitive method in the generation of optical flow datasets, paving the way for advancements in video analysis and processing. Our framework's ability to generate realistic training data from single-view images at scale underscores its potential as a foundational model for optical flow estimation.

The main limitations of our Flow-Anything include that using diffusion for inpainting results in a slightly longer data synthesis time, i.e., about $1.5$s per pair of data. Therefore, multiple processes are required for data synthesis. Another limitation is the relatively limited inclusion of specific perspectives, such as aerial or drone-based views, within our generated datasets. The inclusion of these specialized perspectives could be particularly beneficial for certain tasks where these angles are prevalent.

\ifCLASSOPTIONcaptionsoff
  \newpage
\fi

\bibliographystyle{IEEEtran}
\bibliography{IEEEabrv,egbib.bib}

\begin{IEEEbiography}[{\includegraphics[width=1in,height=1.25in,clip,keepaspectratio]{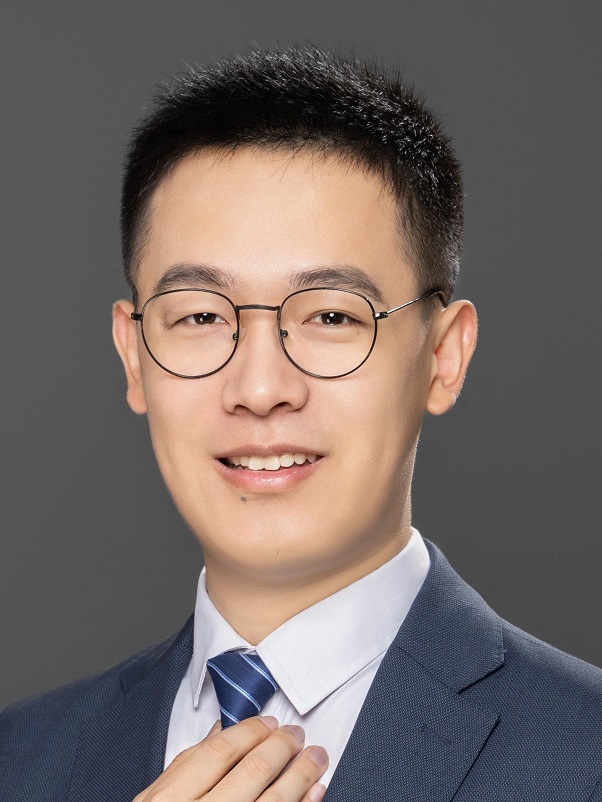}}]{Yingping Liang} received the B.S. degree from the School of Computer Science and Technology, Beijing Institute of Technology in 2022. He is currently working toward the Ph.D. degree in the School of Computer Science and Technology, Beijing Institute of Technology, Beijing. His research interests include deep learning, image processing, and 3d vision.
\end{IEEEbiography}
 
\vspace{-10 mm}

\begin{IEEEbiography}[{\includegraphics[width=1in,height=1.25in,clip,keepaspectratio]{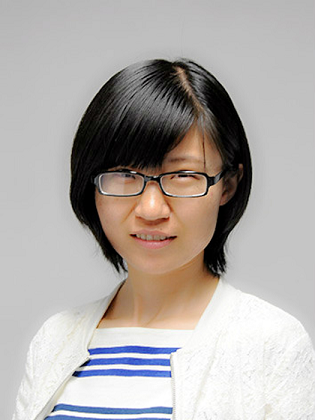}}]{Ying Fu} received the B.S. degree in electronic engineering from Xidian University in 2009, the M.S. degree in automation from Tsinghua University in 2012, and the Ph.D. degree in information science and technology from the University of Tokyo in 2015. She is a Professor at the School of Computer Science and Technology, Beijing Institute of Technology. Her research interests include physics-based vision, image processing, and computational photography.
\end{IEEEbiography}
 
\vspace{-10 mm}

\begin{IEEEbiography}[{\includegraphics[width=1in,height=1.25in,clip,keepaspectratio]{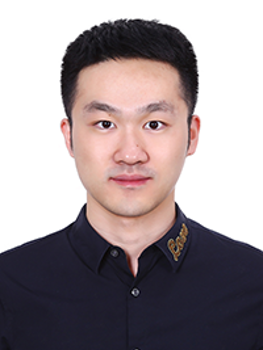}}]{Yutao Hu} received the B.S. degree in electronics
and information engineering from Beihang University in 2017, and the Ph.D. degree in electronics and information engineering from Beihang University in 2022. He is an Associate Professor in the School of Computer Science and Engineering at Southeast University. His research interests include machine learning and computer vision.
\end{IEEEbiography}
 
\vspace{-10 mm}

\begin{IEEEbiography}[{\includegraphics[width=1in,height=1.25in,clip,keepaspectratio]{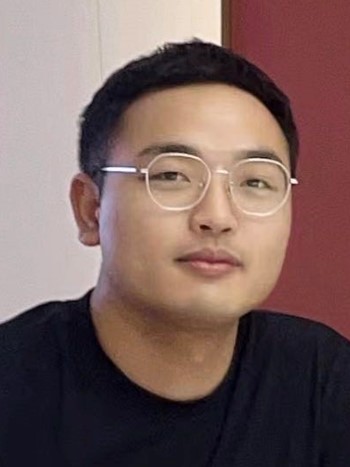}}]{Wenqi Shao} received the B.S. degree in School of Mathematics at University of Electronic Science and Technology of China in 2017, and the Ph.D. degree in Multimedia Lab of the Chinese University of Hong Kong in 2022. He is a Young Scientist at the Shanghai Artificial Intelligence Laboratory. His research interests include multi-modal foundation models, large language model compression, efficient transfer learning, and their applications in multimedia. 
\end{IEEEbiography}
 
\vspace{-10 mm}

\begin{IEEEbiography}[{\includegraphics[width=1in,height=1.25in,clip,keepaspectratio]{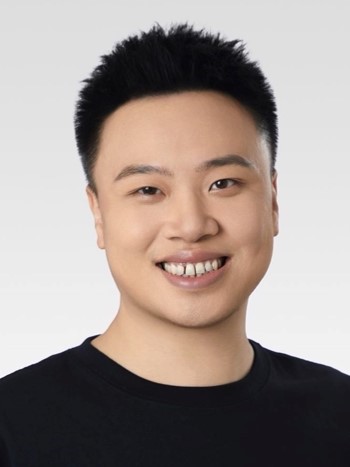}}]{Jiaming Liu} received the B.S. degree in electrical engineering in Shandong University of Technology in 2010, and the Ph.D. degree in pattern recognition and intelligent systems from Tongji University in 2014. He is the Vice President of Artificial Intelligence at Tiamat AI. His research interests encompass generative AI, optical flow, and optical character recognition.
\end{IEEEbiography}
 
\vspace{-10 mm}

\begin{IEEEbiography}[{\includegraphics[width=1in,height=1.25in,clip,keepaspectratio]{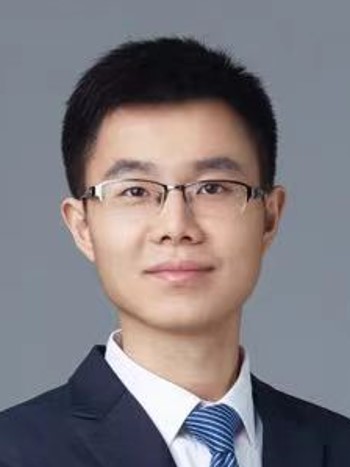}}]{Debing Zhang} received the B.S. degree in math and applied mathematics from Zhejiang University, China, in 2010, and received the Ph.D. degree in computer science at Zhejiang University, China, in 2015. He is currently the leader of LLM research and application at Xiaohongshu. His research interests include machine learning, computer vision, data mining and large language and multimodal models.
\end{IEEEbiography}

\end{document}